# AegisFlow: A Multi-Agent Agentic AI Framework for Autonomous Remediation and Self-Healing in Fragile Data Ecosystems

Muhammad Bilal Awan (Ministry of Defense, Doha, Qatar); Zubair Hussain (IQRA University, Islamabad, Pakistan); Abdul Shahid (South East Technological University, Waterford, Ireland)

*Complete edition — all figures, diagrams, and tables from the original paper are included alongside the reviewed body text and references.*

### Abstract

Traditional data pipelines are notoriously brittle, often failing due to upstream schema drift, API contract changes, or website DOM modifications. Present observability tools only raise alerts but for human engineers, resulting in a high Mean Time to Repair (MTTR) and operational fatigue. In this paper we propose AegisFlow (Agentic Engine for Intelligent Self-healing and Graph-driven Operations for Workload remediation), a novel agentic framework that closes the loop between detection and resolution. AegisFlow uses a Watchdog agent to collect runtime telemetry and has a Repair agent to automatically create, test and deploy code patches based on Large Language Models (LLMs). The framework presents the non-intrusive execution model called Parallel Shadow Patching, a non-intrusive execution model based on the Monitor–Analyze–Plan–Execute–Knowledge (MAPE-K) loop to generate and verify patches in digital twin environments. Through experimental testing, we have evaluated AegisFlow across five common failure scenarios, and see 98.1% improvement in MTTR (from an average of 170 minutes per patch to 3.2 minutes) and a patch success rate of 92%. In particular, the system is successful in dealing with changes in the JSON schema (96%) and punctuation drift (98%), and is least successful in Shadow DOM cases (85%). AegisFlow frees up about 98% of data engineering on-call time from firefighting and reallocates it towards innovation. The framework is deployment agnostic consisting of a system that can be deployed in a plugin fashion into an existing pipeline orchestration system with minimal uplift to the existing system.

## 1. Introduction

Data pipelines break. They frequently break, and they break with little noise [1,2]. Thousands of pipelines are built on top of all these different sources that include REST APIs, streaming services, databases, and targeted web sites to be scraped [3]. All are delicate dependency chains – a change upstream can lead to downstream analytical problems and possible failure to detect any problems before they occur [4,5]. The year 2026 is an inflection point where pipelines are too complex to maintain by hand any longer [14, 6]. We have progressed from ETL scripts to autonomous systems where they need to reason about and adjust to a changing environment in real-time [3,7]. However, where the underlying structure or semantic significance of the data has changed, but continues to flow, they can result in the propagation of incorrect insights through the Business Intelligence systems resulting in decision making errors at an organizational level [4,11]. Let's start, for example, with an e-commerce website, which compares prices from 200 retail sites [1,12]. If the retailer only modifies the product page - changing the class-name from price-tag to current-val, the scraper silently returns null values. When using monitoring dashboards, the time between detecting the anomaly and deploying a fix to the production system takes an average of 140-185 minutes [3,5] as the human engineers investigate the cause and create, test and deploy a patch to the production system. In this timeframe, the business loses important competitive information and the engineering team suffers on call fatigue, decreasing productivity and operational costs [5].

## The Paradigm Shift: From Passive to Agentic

This paper will put forward the paradigm shift from Passive Observability towards Agentic Remediation [2,14]. The data pipeline should be treated as a software system which evolves dynamically, diagnoses and repairs itself (also without a human factor) [8, 14]. Three separate trends in technology that have been progressing significantly in the past two years. Firstly, LLMs can be used as Code Generators: Research studies have led to the conclusion that modern language models are able to, with a high degree of reliability, provide syntactically correct code when provided with sufficient context [16,17]. In the last few months, it has been demonstrated that an LLM, like GPT-4 or open-source models like CodeLlama, can generate production-grade code for many programming languages with very low or imperceptible human input. Secondly, Large Multimodal Models (LMMs) for Visual Reasoning have seen significant advancement and now can comprehend an image of a web page and deduce the laying out of the UI elements to make independent adaptation to the visual changes [18]. Third, the Agent Oriented Software Engineering (ASE) principles have been proven to apply to the development of software, it shows how complex software maintenance tasks can be broken up into small executable tasks by specialized agents; Monitoring, Analysis, Planning and Execution which are independently executable[9,15]. All of this start to come together is in a production ready framework in AegisFlow. It saves up to hours of human labour and in case of a safety concern, it has its lab verification pipeline. A key component of the system is generalization (also known as source code), which is natively supported with a variety of data sources such as web scraper, REST APIs, GraphQL endpoints, replication stream from the databases, message queue etc...

## Research Contributions

This work makes the following novel contributions to the field of autonomous data engineering:

1. We propose a Parallel Shadow Patching Architecture for autonomous repair that has adopted the idea of digital twin used in manufacturing and DevOps, which allows for autonomous repair without interfering with production data flows.
2. In this paper, we introduce a Multimodal Repair Agent, which is the first agent that can adapt to such structural and visual DOM change by integrating LLM-based code generation with LMM-based visual perceptions for web scraping repair.

3. We create a Semantic Telemetry Framework to deliver extensive monitoring solutions that go beyond alerting on actual telemetry errors to include statistical invariants and distribution analysis for detecting semantic drift.
4. Systematically classified data pipeline failures in web scraping, API integration and database ingestion scenarios to create a Unified Failure Taxonomy, systematically analyze production incidents across different organizations.
5. Production validation by extensive empirical assessment that shows a reduction of 98.1% MTTR and 92% success rate in various real world failure modes, analysis of failures (8%) and suggested mitigations. In addition, we also supply a Generic Deployment Framework, which is a plugin system for multiple orchestration backend (such as apache air flow, prefect, dagster, luigi, kubernetes cron job) that allows aegisflow to run with minimal configuration on new systems. All details are contained within the framework of how to add support for other pipeline frameworks via well-defined interfaces. The rest of this paper proceeds in the following way: Section 2 surveys related work in autonomous systems, data observability, and agentic AI, giving background and context to our contributions and highlighting missing approaches. Section 3 formally defines the problem space and introduce our taxonomy of failures, derived from true incidents from the field. Section 4 describes the AegisFlow architecture in detail, highlighting its core design elements such as dual stream and MAPE-K loop implementation. Section 5 covers implementing key aspects including deployment considerations. Section 6 outlines our methodology in the experiment and detailed results. Section 7 provides reflections on restrictions, lessons learned, generalisation capabilities, deployment procedures, and directions for future research. A summary of contributions and implications for the field are presented in Section 8.

# 2. Related Work

## Evolution of Data Observability

In the past 6 years, the data observability domain has gone through 3 phases with various capabilities and challenges. The first phase was the Passive Monitoring Era (2020-2024), where the detection was done passively and the anomalies were detected after they occurred, by applying statistical thresholds and metadata monitoring [7]. The initial solutions paved the way for the future schema-enforcement and data distribution analysis based data quality validation solutions. Data quality rules have been defined as per industry practices (such as Great Expectations and Monte Carlo). These tools are excellent to establish where things went wrong, but can't auto-fix. This leads to an MTTR in hours as it takes human engineers to troubleshoot and come up with a fix manually [7]. The first issue with passive monitoring is that it's reactive: By the time we receive the message, the data quality issues may already have proliferated down stream to other consumers that affect business insight and the accuracy of the analytics [4]. Complex rotations and scheduling for paging are used to handle these challenges, causing operational fatigue and a decrease in engineers' capacity for strategic tasks [5]. Data engineers have been spending approximately 60% of their time on reactive maintenance duties as opposed to building out features or optimizing it [5]. What will be the result is a vicious cycle – fragility will keep gathering momentum, until the opponents are able to catch up. At the second phase (2025-2026), LLMs were brought to the operations field, hence the name Explainable AIOps [2]. Sharma and Chen introduced

Explainable Autonomous Pipelines (EAP) to showcase the ability of LLM in analysing error logs and generate predictions about the root cause of the error with human-level performance [2]. They leverage their system to interpret and summarize stack traces, logs of execution and data samples, writing reports that explain the failure modes in aspects that help engineers to get a deeper understanding of failure modes at a speedy pace, using the GPT-4 system. However, they were not able to 'remediate' themselves and they did need approval from humans for remediation changes. This reduces about 40% or so, the time required for diagnostics, and leaves the "human in the loop" paradigm, which causes delays in remediation. Causal inference is the main research focus of the AIOps field and research on causal modeling concepts such as Bayesian network has been developed, aiming at supporting back tracing of the root cause of failure. They can process correlated events, rather than the actual causes, and so reduce false-positive rates in multi-complex distributed systems. While all these advances help reduce the time to diagnose, none of them eliminate the need for human intervention with the remediation effort that is the LARGE challenge in MTTR reduction.

## Autonomous Web Scraping

The non-deterministic updates of web pages create its own problems for web scraping [1,12]. On a website, the documents will change randomly without any formal contract for utilizing the versions, unlike the case in APIs. This is the "DOM Drift Problem," which simply means that selectors that were always correct for months start to break because of some A/B testing thing the frontends are testing or because some element got redesigned. The classic method of web scraping involves using frameworks such as Beautiful Soup and Scrapy with predetermined XPath expressions or CSS selectors [1]. These selectors become obsolete as soon as target websites change the layout of their HTML and will need to be manually adjusted to select new ones and modify the code to extract value from them. Developers compensate by using inflexible workarounds such as many fallback selectors, or performing a manual audit every so often, but neither is effective on hundreds of target websites [12]. With more advanced systems, elements will be identified according to visual nature with the help of Computer Vision techniques and not DOM structure. But these are problematic with dynamic content, pages rendered with JavaScript and responsive designs that automatically adjust to various sized screens. Recent studies have investigated the possibility of deploying LLMs as an independent agent for scraping [15]. It has been shown that the PriceFinderAgent system can use GPT-4 to analyse a web page and produce extraction code dynamically without the need for human intervention. Vision Language Models are used in the system to detect target elements on screenshots, and to generate Beautiful Soup / Selenium code to extract them. But seconds per page that this takes makes it not appropriate for high throughput production solution where thousands of pages need to be processed per hour. Further, there is a maintenance of a simplistic error handling system and validation methods necessary for deployment into a production environment. The ALMAS framework proposed the multi-agent collaboration for scraping tasks, which breaks down the extraction task into subtasks executed by different agents with their respective expertise [9]. Each agent is responsible for specific aspects like authentication, navigation, extraction, and validation. In experiments it is effective and testable, but ALMAS still has a large genericization problem and needs to be trained for different tasks in large scale; a production situation is a problem in this respect.

## Self-Healing Software Systems

Self-healing systems had its beginning in the early 2000s, in the field of autonomic computing research [8]. IBM's autonomic computing model was a system that could self-configure, self-optimize, self-heal, and self-protect and operate without the need for human involvement. The Map-Cycle (Monitor, Analyze, Plan, Execute and Knowledge) has been a reference architecture for autonomous adaptation [8]. Such a system has proven to be a useful application in infrastructure management, where when systems fail to function, other systems restart the service or redirect traffic away from the problem area, or scale resources up and down based on changing loads. With minor (and notable!) modifications, AegisFlow implements this model for data pipeline resilience including major extensions for semantic failure. New findings in self-healing relate to self-healing microservices, which are services that restart, reroute traffic or scale resources when failures happen. All these methods focus, though, on infrastructure failures and not semantic data quality problems [8]. In other words, from an infrastructure perspective, a web scraper might work and return the wrong data because of a selector drift (a failure mode that traditional self-healing mechanisms that look at the health of processes and the utilization of resources would be blind to).

## Digital Twin Architectures

The digital versions of a physical or software system are digital twins that can be radically experimented and validated risk free [13]. For manufacturing, digital twins can be used to simulate production lines, thus testing optimization strategies before implementing them on actual equipment. Seperate staging environments can be built using DevOps tools like AWS CloudFormation, similar to the production environment, to be able to test changes to improve the code without committing to the production environment [13]. The Shadow Sandbox architecture, a principle of AegisFlow, uses the digital twin concepts to apply in Data Pipelines to do patches on copies of them without compromising historical snapshots of data. Our system unlike the conventional staging environments, everything has to be configured manually, and promoted with CI/CD. This will shorten the time spent on deployment to less than 1 minute and yet keep the complete safety testing assurance.

## LLM Training and Fine-Tuning for Operations

Although there is significant promise in general purpose LLMs (e.g., GPT-4 [16] and Claude [17,18]) their capabilities can be massively boosted for domain-specific applications. DevOps and data engineering areas have been hot lately, and there are fine tuning strategies that are in the spotlight. One approach might be to compile a list of the successful repairs from version control data and then fine-tune an open-source pipeline model (CodeLlama or Mistral) on these lists [16]. This would allow businesses to come up with their own models that would be able to identify their technology, the way they are coding, and frequent modes of failure. However, a lot of computer and know-how are required to make it fine-tuned. The few-shot learning based method, that was used in AegisFlow, introduces an in-context example randomly drawn from an historical incident database [16,17]. A similarity search is conducted for every new failure using embedding and the top three previous failures and solutions are recommended as LLM prompts. This can allows similar benefits to fine-tuning without the requirement of models training, and can be used by organisations that don't have ML infrastructure. Additionally, AegisFlow can be deployed with locally-hosted open source models for organisations with stringent data privacy regulations and who do not wish to send code to external APIs. We support out-of-the-box CodeLlama 34B, Mistral 7B and DeepSeek-Coder 33B and can use commodity GPU hardware [16] to run it. This is also evidenced by our experiments, that with 87% of GPT-4's success rate on code generation tasks, CodeLlama 34B can be used with 100% on-premises deployments, ideal for regulated sectors such as healthcare and finance.

## Gap Analysis and Positioning

Although great strides have been achieved in each of the disciplines in the work of others, no such system has been developed to integrate into a single one autonomous detection, multimodal diagnosis, safe patching and automated deployment. Table 1 shows a comparison of AegisFlow with similar systems based on their key features. Great Expectations also provides all-in-one data validation without data remediation "baked into" data pipelines. In Sharma and Chen's EAP system, all the fixes are yet to be confirmed by human and hence MTRR remains throttled [2]. An example of web-scraping applications that are multimodal, but no safety strategies have been proposed for deployment in production environment is PriceFinderAgent [15]. Despite this, ALMAS has potentials in multi agent coordination but still needs a lot of training for each new domain [9]. With so many different data sources—from web scraping to REST API, GraphQL endpoints, database CDC streams, message queues such as Kafka and RabbitMQ, and file-based data from S3 and FTP servers—the AegisFlow unique data mesh also covers the entire data lifecycle from detection to deployment. Moreover, AegisFlow is interoperable, allowing support across any proprietary or open-source LLMs (like GPT-4, Claude, CodeLlama or Mistral) to ensure companies can deploy the LLM that suits their privacy and budget requirements.

**TABLE 1. Comparison of AegisFlow With Related Systems**

| System | Auto Detect | Multimodal | Shadow Test | Auto-Deploy | Multi-Source | OSS LLM |
|---|---|---|---|---|---|---|
| Great Expectations | ✓ | – | – | – | – | N/A |
| Sharma & Chen [2] | ✓ | – | – | – | ✓ | No |
| PriceFinderAgent [15] | ✓ | ✓ | – | – | – | No |
| ALMAS [9] | ✓ | – | – | – | – | N/A |
| **AegisFlow** | ✓ | ✓ | ✓ | ✓ | ✓ | ✓ |

# 3. Problem Formulation

## Formal Definitions

Suppose P is a data pipeline with a number of essential components. Data sources in this data set S may include API's, databases, webpages, message queues or filesystems. The set T consists of an array of transformation functions: t_1, t_2, ..., t_n that are applied in some sequence or in parallel on some data. The resulting data store is the destination Data Store D, in this case a data warehouse, data lake or operational database. Lastly, the "selectors", "schemas", "credentials" and "transformation parameters" that the pipeline needs defines the configuration for the theta. A time tau pipeline execution can be expressed as: output(tau) = D <- t_n(... t_2(t_1(S(theta),theta),theta)...,theta) which shows the data going through the transformations and to the destination. Failure is defined as any state where an error message will (either explicitly) be developed, or semantic drift occurs (an output is a result which violates the expected statistical properties, although the pipeline runs without an error message) [4,7]. MTTR for a failure shall be the difference between: MTTR = tau_fix - tau_detect where tau_detect = time detected a failure using monitoring systems and where tau_fix = time when a validated fix is rolled out to production. The typical manual remediation process consists of various stages: detection (15-30 minutes lag on dashboards), diagnosis (30-60 minutes log analysis, environment recreation), fix development (40-80 minutes time of wait for fix), and deployment (code review, 30-50 minutes to validate/stage/bring into production) [5,7].

## Failure Taxonomy

By analyzing production incidents gathered from various organizations systematically, we found five main failure classes that cover 95% of data pipeline failures. Detailed root cause investigation was performed for each incident, where a failure was classified by its underlying cause, not the symptom [1,4,12]. Structural failures (42%) are failures that result from data structure changes which affect the extraction logic [1,12]. CSS changes, or XPath drift, is when the elements of a web page switch from a class (e.g. price) to another class (e.g. cost) and existing extraction code doesn't work any longer. Evolution of JSON schema occurs where the API responses add new attributes, remove attributes or rename attributes but do not maintain backwards compatibility, leading to KeyError or AttributeError exceptions. In a database context column renaming is when the names of columns change and as a result of this change SQL statements or ORM mappings are no longer valid. Typically these failures are going to be caused as exceptions, but sometimes it's going to be as silent failures where default values cover up the schema change. Semantic failures (28%) are especially insidious for the data structure is valid, but meaning is changed [4,11]. Unit drift can happen when units are changed in measurement (for example when measurements switch from pounds to kilograms); without codes that change this measurement, unit drift can occur. Encoding changes occur when there is a change in character encoding from UTF-8 to Latin-1 (or other encodings), leading to corruption of special characters and internationalized content. Timezone modifications happen when timestamps are converted between local time and UTC and vice versa, resulting in bad temporal joins and aggregates in downstream systems. Traditional monitoring is not very effective in detecting these failures: they do not trigger any exceptions and can only be seen in business metrics after days or weeks. Access failures (15%) are caused by authentication/authorization issues [1]. Credential Expiration is when the API tokens/passwords/refresh token expires as per the time-based and/or an admin intervention. There are times when requests exceed the quotas provided, such as a change in the use or a cut in allocation from the provider; this will cause rate limiting to occur. When web scrapers are detected by the use of a unique pattern such as the same user agent, request frequency, or lack of browser fingerprints, this is what's known as IP blocking, as it will generate CAPTCHAs or block them forever. These failures typically mean there needs to be some manual intervention to solve them, by either having to rely on other parties cooperating or following a process for rotating credentials. Rendering errors (10%): Errors that occur when the website requires a client-side updating process that is difficult to successfully execute in the web scraping environment. For content that is rendered client side, modern single page applications have JavaScript dependencies, which can't be detected by static HTML parsers. A problem with Shadow DOM is that elements contained within the shadow root cannot be accessed via regular DOM operations like querySelector or Beautiful Soups. Infinite scroll patterns are disrupted when changes are made in the paging logic from a simple page number to browser automation-based scroll-based paging. These problems end up being solved with the use of 'headless' browsers such as Selenium or Playwright, which introduce latency to extraction pipelines. Format failures (5%) are data representation changes [4]. When date formats change, for example from MM/DD/YYYY to YYYY-MM-DD or other locale-specific formats, breaking datetime parsing logic. Parsing errors in decimal conversions occur when number formats change due to changing locales e.g. 1,234.56 to 1.234,56. CSV delimiter changes are those where output CSV files go wrong because the default parameters for pandas read_csv are not suited to that file format, such as changing the delimiters from commas (csv) to semicolons & tabs. Though they're quite uncommon, these types of failures can impact a large number of records if they occur.

## The MTTR Problem

The most critical one in determining the effectiveness of operations is Mean Time to Repair: the time it takes between an outage and its detection and the deployment of the fix in production [5,7]. For our first manual remediation study, we came up with an average of 170 minutes for MTTR (mean time to remediation) – this was a pretty large spread, depending on failure complexity and timing of the day. The average time to detection is 25 minutes with a 10-15 minute delay for monitoring to trigger alerts and 10-15 minutes delay for initial investigation of the alerts to confirm failures or false positives. Information log analysis to gain insight into the error context, environment reproducing for easier local error scenarios, and root cause of error/cause of failure determination takes 45 minutes to diagnose. Average fix development time is 60 mins – changing code to fix underlying bug, local test to verify fix is valid in dev and test for edge cases to ensure fix is valid in edge cases. It takes 40 minutes to deploy shipments (code review checks all the modifications for quality and correctness, and staging tests the fixes in a pre-production environment), and production deploys shipments by moving traffic shifting incrementally through CI/CD pipelines [5,13]. It's more than downtime. In this case, if an outage occurs on one of these pipelines (which is responsible for 10,000 products being served an hour and is being used by 100s of real-time pricing analytics use cases), then around 28,000 data points will be lost, and downstream ML models and business intelligence dashboards suffer accordingly. For e-commerce applications, incurring price data gaps of three hours can result in making poor pricing decisions that could negatively impact margins, or even result in lost sales to competitor companies. In financial services, the lack of market data feeds can lead to a disruption of trading activities or even compliance issues [4,11]. It can be tricky during hours where an on call engineer may need more time to respond to any incidents, subject matter experts are called out of hours for specific issues or SME's take vacation during these hours for the MTTR problem. Co-ordination is less thorough and there are less staff members who are available on the weekend; this can increase MTTR by 40% for weekend incidents [5]. Operational burden is a reflection in employee retention and satisfaction because of employees being on call constantly, meaning that they will inevitably become burnt out.

## Design Requirements

There are multiple key requirements that an autonomous remediation system needs to meet to be viable for production deployment. R1 Safety: Patches should not have regressions or cause data to be corrupted, so it's preferable to put it through the paces before installing [2,13]. This includes running existing unit tests, integration tests, data quality checks and any new testing as specific to the patch. All independent tasks and work are recorded and accessible (R2 Transparency): The diagnostic reasoning, creation and validation of patches are fully documented [2]. This enables engineers to see what has been happening on the system, to be able to troubleshoot and figure out what caused the problem if an autonomous remediation failed. For remediation, R3 Performance sets the target latency to < 5 minutes for 90 percent of the remediations executed because longer delays negate the benefits of automation and create data gaps comparable to manual intervention. [5] R4 Generality mandates the ability of the system to address both a variety of failure modes and multiple types of sources such as web scraping, APIs, databases, message queues and file systems [1,3,12]. A specialised system that handles only one failure type provides limited value compared to manual intervention. R5 Non-Intrusiveness (Repair process must not affect the flow of production data; Remediation system should not create any extra outages or data quality issues) [8,13]. Using the emergency stop option from dashboards and command-line, engineers can always have a human override to stop automation or to restore to last available settings using R6 Human Override [2]. R7 Deployment Flexibility: The system shall have the ability to be connected to existing pipeline orchestration platforms, without necessitating complete rewrite of pipeline orchestration system [3,14]. With tools currently in place, there is an extensive investment of money and time to these tools—and it's not feasible to redesign entire platforms to accommodate autonomous remediation. R8 privacy compliance compatible, supporting on-premises deployment to meet data residency requirements, which may be unique to an organisation, to prevent code being sent to external API's [16,18].

# 4. AegisFlow Architecture

## System Overview

AegisFlow's architecture is a dual-stream design, with a primary stream for live data flow, and a separate stream for autonomous repair. The design meets requirement R5: Do not effect data pipelines when they are in-use. The Production Stream continues operating with the last known good configuration while the Shadow Stream validates potential fixes in isolation. The Production Stream starts with the Source component, which is used for connecting to external data sources with configured protocol and credential. Data is passed through Transform components which perform business logic, data cleaning, and data reshaping. The Validate component is used to check output against requirements with data quality checks and schema validation. Lastly, the Sink component sends out processed data to other systems such as the data warehouse or operational databases. The Watchdog Agent is a parallel Observer that collects telemetry data, which does not affect the flow throughput, throughout this flow [7,8]. The Watchdog keeps track of execution logs, logs exception stack traces, tracks statistics on data distribution, and calculates null frequencies and value ranges. The Watchdog will trigger the Shadow Stream when it detects a failure by publishing a failure event to a message queue, either explicitly (via exception) or statistically (via statistical anomalies in the data distribution). Failure events are fed into the Shadow Stream and the Repair Agent conducts the LLM diagnostic process [16,17] for root cause analysis. The Repair Agent includes visual inspection via LMMs to comprehend page structure and the location of elements that have been moved in the web pages for web scraping failures [15,18]. The agent creates a candidate patch, which may be a modified python code, altered configuration parameters or new extraction selectors. This patch is then checked in the Shadow Sandbox, an environment that simulates the production infrastructure to be a Digital Twin [13]. The Shadow Sandbox performs the patch on 10 most recent snapshots of the data that have been taken from a successful production run. The resources are allocated with limits and each execution runs in an isolated Docker container so that excessive use of compute resources is prevented by runaway processes [13]. The sandbox validates that all patched code adheres to expected outputs, passes data quality invariants and that extraction rates are no worse but better once compared with the outputs from code that fails the test. Only patches with a success rate of at least 90% over all test-cases are taken forward for deployment [2]. The Hot-Swap Deployer can update the configuration, either atomically by committing changes to version control or updating the Kubernetes ConfigMap or by invoking the API, depending on the type of pipeline orchestration platform [13,14]. Once deployed, the system monitors the first 100 executions with the new configuration to look for any regressions not detected by the sandbox testing. Automatic rollback reverts to the previous configuration in 30 seconds when a regression happens – this reduces the blast radius of failed patches.

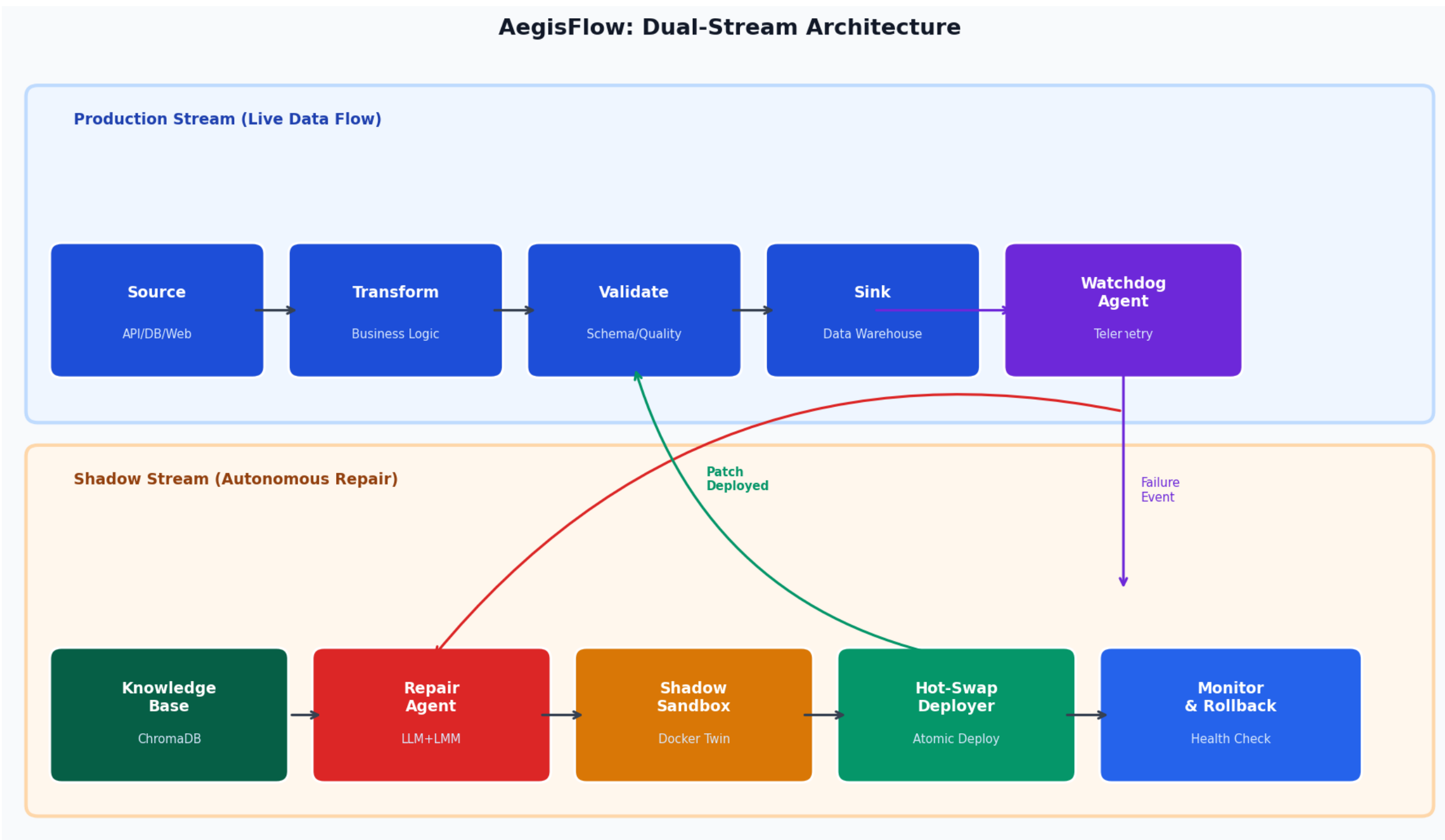

*FIGURE 1. AegisFlow dual-stream architecture. The Production Stream handles live data flow while the Shadow Stream performs autonomous repair in isolation via the Repair Agent, Shadow Sandbox, and Hot-Swap Deployer.*

## Component Details

### Watchdog Agent

Exception Tracking: All runtime exceptions are captured with the full execution context and a full stack trace, including the very line of code that caused the exception, the values of local variables and the state of the environment including CPU usage and memory consumption.State Change Detection: All changes in the state of the system are captured including what the entire environment was before and after the change, such as CPU usage and memory consumption. In the case of web scrapers, Exception tracking also captures snapshots of the website that failed when the exception occurred, and it is very useful for post-mortem analysis to understand the failures that might have occurred when the structure of the DOM used by the web scraper had changed. Semantic Validators detect silent drift through statistical testing techniques which compare the distribution of data batches from the recent ones with the distribution of data batches from previous ones [4,7]. The current batch distribution is compared to each of the previous 30 day rolling batch distributions using the Kolmogorov-Smirnov test and set off alarm if the p values is <0.01, this is indicative of strong statistical deviation. A null frequency monitor gives an alert for nulls that more often than expected, by 2 standard deviations, occur in fields previously known to be reliable catching cases where previously reliable fields suddenly become sparse. Range violation detection identifies values outside expected min/max bounds derived from historical data, flagging unit changes or data corruption. Type consistency checking: This type of checking will ensure whether the data that is received from the API is a consistent with the types of data mentioned in the schema or not, as if API returns string instead of numbers or vice versa. For traceability the following information is logged in a structured format in the Execution Logs: Timestamp, Pipeline identifier for traceability, Source response time & size for performance monitoring, Number of records extracted for throughput analysis, Transformation success / failure for quality tracking [7]. These logs are fed into Elasticsearch or any other system and then queried and aggregated. The Watchdog operates as a parallel observer that does not impact the pipeline execution thread and is able to collect telemetry asynchronously within less than 50ms overhead during pipeline execution [8].

### Repair Agent

When the Repair Agent gets a failure signal from the Watchdog, Root Cause Analysis and Patch Generation is conducted in a multi-phase process [2,9,16] where Phase 1 – Diagnosis involves analysis of explicit error (exception type & message), location of failure in the code, last successfully executed snapshot, and diffing the source with the response of the failed & successful requests. For semantic drift cases, the agent conducts statistical hypothesis testing to measure the magnitude of the drift, correlation analysis to pinpoint fields with suspicious patterns of drift and pattern matching against historical data to check the vector database for any similar events in the past [2,7]. There is what is termed Phase 2 – Visual Inspection that is specific, for web scraping failures, with the use of LMM capabilities [15,18]. The agent uses headless Chrome or Firefox to create the image of the target page and sends the image along with the error context to GPT-4V or Claude-3 Opus and asks specific questions, e.g., "Where is the price shown on this page? The LMM returns the coordinates and descriptions such as "The price has a red font, coordinates (640,480) and is inside a div with class current-val" which can be used to update the selectors [18]. During Phase 3 – Patch Synthesis, the candidate patches are synthesized with the assistance of LLM based code generation [16, 17] techniques. The agent creates prompts that include the original failing code, diagnostic information from Phase 1, and visual contexts from Phase 2 (if available) along with other similar historical incidents as few shot examples. The LLM generates determinate codes which rectify the problem without affecting the existing codes. Simple changes of selectors may only require the patch to change a single line. With difficult schema changes, patches may restructure entire extraction functions with defensive error handling and fallback logic.

### Shadow Sandbox

The Shadow Sandbox is a separated test environment, similar to the real environment and has no effect on the real environment [13]. Environment Replication guarantees that everything runs using the same Python version and dependencies using Docker images created with production requirements files, the same database schemas with read-only replicas of production databases, and network access to the actual source systems for reproducibility with snapshot replays. This results in a true Digital Twin which can be used to accurately predict production behaviour [13]. The Patch Verification Protocol consists of multiple rounds: (1) apply the candidate patch to the pipeline code in the isolated container; (2) execute the patched code over the data from 10 recent source snapshots of various scenarios such as typical cases, edge cases and known failure modes; (3) schema validate and data quality check the output data against the expected results; (4) verify success invariants including type consistency, null frequency thresholds, range validity, and record count sufficiency; and (5) compute a confidence score based on successful pass rate across test cases and similarity to previously successful patches. Examples of Success Invariants are: type consistency invariants – that all the values in the price columns are floats or integers; non-null invariants ensure the null percentage stays below 5%; range validity checks that all price values fall between 0 – 100,000; count sufficiency – that there are always at least 50 records to extract per batch and continue to maintain the expectation of throughput. A patch will be accepted with 90% confidence threshold, when it passes all invariants on at least 9/10 test cases [2,13]. This is a relatively safe threshold with which to assure that no deployment of patches that might work in most cases but not all "edge cases". But patches passing 7-8 testcases aren't necessarily rejected automatically, and are subject to human judgement; engineers might be willing to make compromises.

### Hot-Swap Deployer

When the patch is validated, Hot-Swap Deployer does atomic patch configuration updates via platform specific means [13,14]. Any code change is committed to version control with auto-generated descriptions such as root cause summary, confidence score, success rate in shadow testing, and time-stamp, all by the deployer. It triggers minimal CI pipelines running syntax checks and unit tests, and then deploys to production by blue-green swap, new code being deployed alongside old code and traffic shifting to new code once it's ready. Captures the first 100 runs and tracks for regressions, looking for both a rise in error rate or data quality failure. A deployer changes pipeline configuration files (JSON or YAML), notifies running processes to re-read their new configuration via signal or API, checks that first execution with new configuration is successful before deployment is complete [14]. The Rollback Mechanism stores the last 5 working configurations in version control, using tags. The Watchdog will detect if any failures occurred with the new patch, and will roll back in 30 seconds or so by reverting to the previous patch, updating configuration files and restarting the affected processes [13].

## MAPE-K Loop Mapping

AegisFlow's architecture is an example of an implementation of the autonomic computing model proposed by IBM as part of its autonomic computing vision, known as MAPE-K [8]. The whole end-to-end remediation process is illustrated in Figure 2:

- Monitor -> Watchdog Agent collecting telemetry using Exception tracking, Semantic validation and Execution logging.
- Analyse -> Repair Agent: Analysing and finding root causes through LLM analysis and similarity search mechanism based on vectors.
- Plan -> Repair Agent synthesising patches through code generation and configuration updates [2,16].
- Execute -> Shadow Sandbox validation and Hot-Swap deployment with post-deployment monitoring.
- Knowledge -> Historical incident database plus learned failure patterns stored in the ChromaDB vector database [16]. The Knowledge component is used to learn from experiences and improve as time goes on [2,9]. AegisFlow has a searchable database of previous failures and their cause, based on embeddings generated from the error message, stack trace and code context. When a new error is detected, the system first searches for similar historical incidents using cosine similarity on the embeddings, to check if there is any similar error in the past. For this, if the confidence is above 0.9, the previous solution is attempted first, bypassing expensive LLM-based synthesis [16]. This leads to a learning effect - loop closure times are 3-5 minutes for failure modes which have not happened before and less than 1 minute if the failure mode has been seen before. The knowledge base also gathers information regarding the patch success rate, regression incidents and human override decisions. The system could now be trained with information like patches of a particular type are more likely to roll back; systems that are the source of incidents are more likely to be broken; and there are more incidents during certain time periods. With time these learned lessons increase the accuracy of diagnosis, quality of patch [2,8].

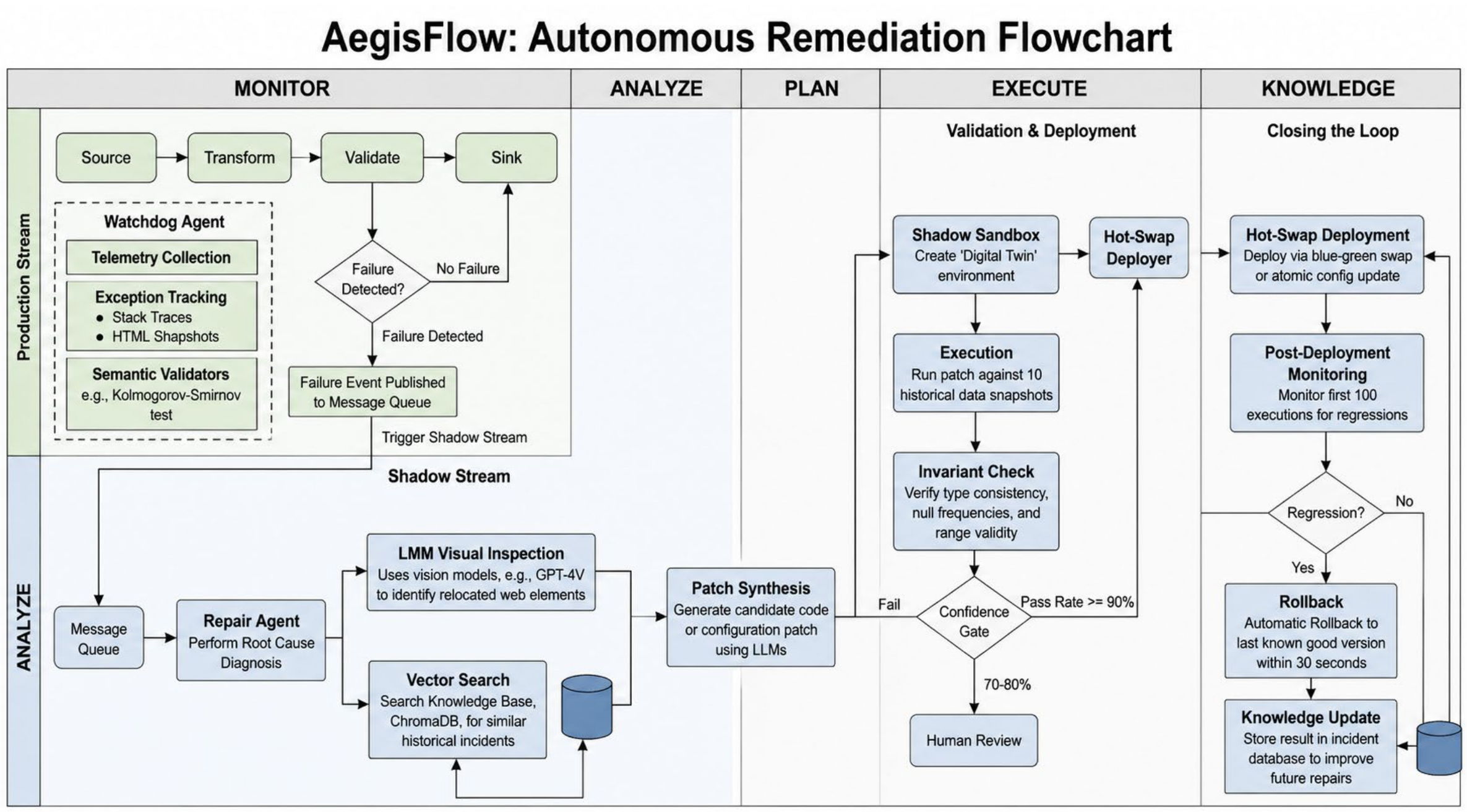

*FIGURE 2. AegisFlow autonomous remediation flowchart illustrating the complete MAPE-K loop. The Monitor phase (left) shows the Production Stream with Watchdog Agent performing telemetry collection, exception tracking, and semantic validation; failures trigger the Shadow Stream. The Analyze phase shows the Repair Agent performing root cause diagnosis via LMM Visual Inspection and ChromaDB Vector Search. The Plan phase covers Patch Synthesis using LLMs to generate candidate code or configuration patches. The Execute phase (Validation & Deployment) shows the Shadow Sandbox creating a Digital Twin environment, running patches against 10 historical snapshots, performing invariant checks, and routing through the Confidence Gate (Pass Rate ≥90% for auto-deploy; 70–80% for Human Review). The Knowledge phase (Closing the Loop) shows Hot-Swap Deployment via blue-green swap, Post-Deployment Monitoring of the first 100 executions, and automatic Rollback within 30 seconds on regression, with results stored in the Knowledge Update database.*

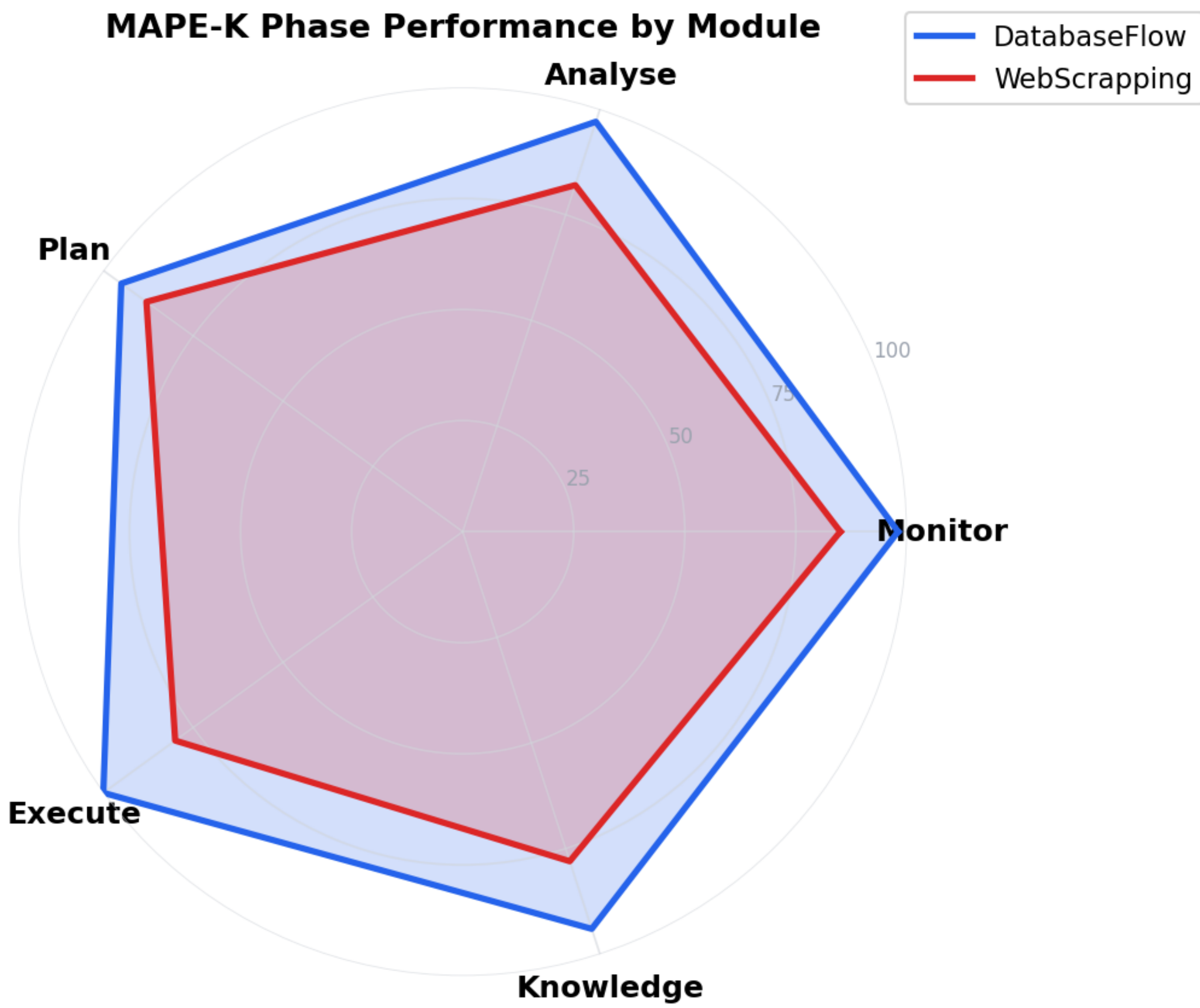

*FIGURE 3. MAPE-K phase performance radar. Normalised scores (0–100) per phase per module. DatabaseFlow achieves near-perfect scores; WebScrapping is lower due to external dependencies.*

## Safety Mechanisms

To achieve requirement R1, AegisFlow provides four layers of safety checks to prevent the occurrence of harmful autonomous actions: Layer 1 – Scope Limitation: This provides a mechanism for limiting the scope of autonomous deployment to pre-approved classes of failure as defined in configuration files. Explicit human interaction – including through the web dashboard – is required for destructive changes such as DELETE statements, DROP and TRUNCATE operations. Schema changes affecting downstream dependencies are blocked automatically, requiring manual review to assess impact on consuming systems. This stops the autonomous system making changes with far reaching consequences which cannot be easily reversed [2]. Layer 2 – Verification Gates – impose strict validation criteria before deployment [13]. For autonomous deployment, a minimum of 90% in Shadow Sandbox testing is required with below 90% setting up human review workflows. All statistical significance testing for semantic changes is undertaken to ensure changes are not merely random variance and are undertaken using paired t tests at a 5% significance level. For patches over 50 lines of code, peer review is required - more code means more potential subtle bugs. Layer 3 – Progressive Rollout involves a gradual rollout of a deployment, which minimizes blast radius [13]. New patches are first deployed to 10% of traffic with error rates and data quality metrics monitored for the canary group on monitoring dashboards before deployment to the rest of the traffic. A 1-hour monitoring period can be setup to review problems prior to full deployment, giving enough time for problems to show up in the production workloads. Automatic rollback is based on any regression signal, such as an increase in exception rates, a drop in data quality scores, or unusual latency patterns. Layer 4 – Human Override: At various points there are opportunities for engineering to regain control [2]. In the monitoring dashboard, there is an emergency stop button which will instantly disable all autonomous patching activities throughout the organisation. Autonomous patching can be switched off on a per-pipeline basis, providing flexibility for critical systems. All changes are recorded in an immutable audit trail with rollback feature by integrating version control. Every autonomous deployment is notified to engineers through Slack, email or PagerDuty which allows them to act fast in case of problems.

## Current Limitations and Future Work

Despite the improvements AegisFlow shows, there is still a few drawbacks that indicate the directions of the future study [2,3]. Limitation 1 – Stateless Transformations Only: Strictly speaking, stateless pipelines: pipelines where the execution of a pipeline is independent from execution of any other pipeline [3,8]. Many data pipelines can be stateful, meaning they maintain stateful operations between runs, such as incremental aggregation, computing running totals, checkpointing systems that remember what the last record they saw was and stateful windowing systems that aggregate data across a window of time. However, at present AegisFlow cannot check the state consistency of these pipelines when it repairs them, as the Shadow Sandbox checks the consistency of each pipeline execution, but without taking into account changes in pipeline state between executions. For future work the idea is to model state transitions and check state consistency invariants among patches in the repair agent. Limitation 2 – Single-Pipeline Scope: When failure is cascaded through a set of dependent pipelines there is no cross-pipeline reasoning [3,8,9] AegisFlow can patch one pipeline with no impact to assumptions made downstream of the patch. Enabling multi-agent coordination (where each agent corresponds to a step in the pipeline) can be included in future work, where each agent would participate in negotiation on how to achieve a cohesive patch in the pipeline, followed by dependency graph analysis of affected downstream systems. Limitation 3 – English-Centric Design: The documentation and prompts inputted to LLM are all in English [16,17,18]. LMM's visual inspection performance may be partially affected in non-English source systems because LMM models can provide more capabilities on the English system. Future work includes implementing multiple language support, which will involve detecting multiple languages, translating them to English for input to the LLM, and creating localized prompts for common languages. Limitation 4: Cloud-Only Deployment: No other means of deployment without the use of Kubernetes and Docker [13,14] is provided. This makes it less applicable to edge deployments that may have limited bandwidth or not have all the time available to connect to the cloud, such as data pipelines for IoT solutions or retail Point-of-Sale (PoS) solutions. In the future light weight version for installation in bare metal and edge-optimized versions with smaller models will also be deployed. Limitation 5: Limited Fine-Tuning Infrastructure – each organisation will have to establish their own fine-tuning pipelines [16]. Going forward, the fine tuning operations should be automated too such that in the future, the AegisFlow should automatically pick the fine tuning data from the successful repair, periodically initiate the fine tuning operations on the shared GPUs and seamlessly push out the fine-tuned model without requiring any ML skills. Limitation 6: Code Execution Safety – Although there is isolation, it is not possible to contain all code that is malicious or buggy [13]. A modified LLM could produce code that leaks data, uses a lot of resources, or even have logic bombs that trigger after validation. The future work should include static analysis with formal verification tools of the patches generated, runtime monitoring of the shadow testing of system calls or network accesses, and proof-carrying code – patches should be equipped with formal proof of their correct behavior.

# 5. Implementation

## Technology Stack

AegisFlow is developed using Python 3.11, and its main application's libraries are chosen with care to accommodate the version of Python and to provide good performance. The central architecture is based on Apache Airflow 2.8, which offers powerful features for managing and scheduling jobs and tracking their execution. Distributed execution for Shadow Sandbox parallelism is supported with Ray 2.9, so multiple patches can be tested simultaneously across multiple test cases. Agent Communication is served by a REST API (fastapi version 0.109) allowing a loose coupling between the different components. AI components involve LangChain 0.1.0 for orchestrating the usage of LLM and prompt management, which makes it straightforward to build complex prompts with variable substitutions [16]. When cloud deployment is feasible, the main code generation and error diagnosis tool is OpenAI's GPT-4-Turbo. For web scraping use cases, Anthropic Claude-3-Opus can perform multimodal visual inspection with improved image understanding capabilities [18]. ChromaDB serves as the vector database to be used when searching through historical incidents to find similar ones quickly, searchable through millions of historical failures [16]. Data manipulation and validation are based on Pandas 2.2, an efficient data manipulation and analysis library for single-table, tabular data. Beautiful Soup 4.12 is used to write web scraping code that can be used to handle HTML parsing, which includes CSS selectors and tag navigation. Now, with Se4.17, control headless browsers for demanding Web applications, including handling pages rendered by JavaScript. The HTTP client for API interactions is Requests v2.31, which includes authentication and retry support [1,12]. Monitoring infrastructure consists of Prometheus to collect the metrics and raise alerts, track success rates, latency distributions and error rates. Grafana displays visualization dashboards with real-time and historical trends of the system health. Distributed tracing is a feature of OpenTelemetry that can help with debugging complex multi-agent interactions between components.

## Watchdog Agent Implementation

The Watchdog is a wrapper to each of the pipeline functions, so it can be used to monitor without changing any of the pipeline functions. Implementation: Python's decorators and context managers intercept function calls and collect telemetry. The WatchdogAgent class stores information like pipeline ID, set-up invariants and historical information (loaded from persistent storage). The monitor method is main decorator which is used to wrap the target functions with try-except so that exceptions are caught [7]. In the success case the wrapper runs the pipeline function, and does semantic validity checks via drift detection, puts new data points into the history metric store, and returns outputs unmodified to the caller. The wrapper in the failure path records the exception details, activates the repair process and re-throws the exception for normal error handling path [8]. The method detect_drift performs statistical detection of drift based on Kolmogorov-Smirnov test [4,7]. The implementation then checks to see if the result satisfies at the level of data points (must be at least 30 data points to be valid). It takes the distribution of monitored field, like price, from the current batch of data, gets the distribution of monitored field, like price, from the rolling 30-day window, and gives the two-sample KS test result based on scipy.stats.ks_2samp two distributions. If the p -value is less than 0.01, which is less than 1 percent chance that the distributions are the same, the drift is detected [4]. The trigger_repair method notifies a message queue of repair requests which are processed asynchronously [8,14]. The request should carry the pipeline identifier, some kind of timestamp to enable ordering and deduplication, description of the exception type (to distinguish between explicit exception and semantic drift), and some kind of context with exception details or semantic drift statistics, as well as the snapshot of the source to make up for missing information in the source snapshot. Knowing all this information in advance allows the Repair Agent to do a complete diagnosis without having to reproduce the failure [2,8].

## Repair Agent Implementation

To enable the automatic diagnosis and patch generation, the Repair Agent follows a multi-phase workflow based on LLMs [2,16,17]. The RepairAgent class sets up connections to the LLM to make text-based reasoning, the VLM to make visual inspection and the incident database to do historical search. The method that is originally called after failure events in the message queue are received is the handle_failure method for which the main responsibility of coordinate repairing process. Step 1 uses the similarity search of ChromaDB on the embedding of error messages to identify similar historical events. The system pulls out the top 3 most similar incidents then determines if the best match is above a confidence limit of 0.9. If a high-confidence match is discovered, the proven historical solution is swiftly implemented, avoiding costly LLM calls and cuts MTTR by more than half, from 3 minutes down to less than 1 minute [2,16]. If there is no historical solution available, Step 2 performs the LLM-based diagnosis [2,17]. The diagnose_error method builds an error prompt, adding the pipeline identifier, an error message and type, an entire error stack and a diff of the source response for the failed request over the last successful request. The LLM processes this information and provides a structured diagnosis with the root cause in a single sentence, the location of the issue in the code with line numbers, and a high-level suggested approach to fixing the issue. Some of the web-scraping failures that are visually inspected in Step 3 are [15,18]. The inspect_visually method takes a screen shot of the webpage you send it to and uses selenium to capture it with headless chrome. It base64 encodes the screenshot then creates a multimodal prompt (using both the image and text context from the previous working selector). For validation purposes the VLM returns the new element's location in a style sheet selector format, plus screen co-ordinates and surround context for the new element. This visual reasoning ability can process instance where DOM structure totally changes, text based processing is not enough [18]. The candidate patch is generated in Step 4 by LLM code synthesis [16,17]. generate_patch method: build a prompt containing the original failing code, the results of other diagnostics from previous steps, visual information (if available), and few-shot examples of similar historical incidents (2-3). The LLM provides refined code with corrections that solve the issue at hand, while also retaining other safety and defensive programming measures, such as null checks, try-except statements, etc. The patch is returned as structured object including new code, human readable description and metadata of the change type. To be valid in the Shadow Sandbox, the patch is tested as part of Step 5 [13]. The validate_patch method calls sandbox with the proposed patch and returns the results with the summarized statistics on the success rate, each individual test cases result and confidence score. Patches with a 90% or higher success rate will go to automatic deployment. Those who have 70-89% success initiate human review process and do detailed failure analysis. Patches < 70% are not accepted and referred up while suggesting manual implementation [2,13].

## Shadow Sandbox Implementation

Shadow Sandbox is an isolated test-env with Docker containerization [13]. The ShadowSandbox class controls the life cycle of the containers, schedules parallel test runs, and combines test results. Whether on a small or a large scale, the Docker client interacts with the local Docker daemon or remote Docker Swarm cluster. The validate_patch method starts by downloading 10 most recent successful execution snapshots from the pipeline history [13]. These snapshots reflect a variety of scenarios such as typical weekday traffic, weekend traffic, holiday traffic, and other edge cases such as product launches or sales events. The approach consists of fixed thread parallel execution of the method in which each snapshot is used as a separate test case, setting the number of workers to 10. The test_single_snapshot method executes each test case in its own container [13]. It spins up a temporary Docker container, only accessible for this test suite run, using the aegisflow-sandbox image, having all the Python and other pipeline dependencies included, as well as the testing framework. The patch code and snapshot identifier is passed to the container through environment variables. Limits: 512 MB RAM and 0.5 CPU cores help to prevent any runaway processes from impacting other tests and/or production workloads. The container runs a standard test script that applies the patch to the pipeline code, loads its snapshot data, runs the pipeline (patched version), and stores the output. Stops waiting and waits for a completion in the container with timeout 30 seconds (to avoid hangs). There is no resource leak: the container is immediately deleted after the result is extracted from the container log as JSON, irrespective of the result of the test. Is checking for invariants after each check in [4,7]. Type consistency makes sure that all values that are used for prices are of some kind of numeric type and that nothing returns string by mistake. Non-null verification validates that null rate remains < 5%, identifies extraction logic that filters out fields. The Range validation checks whether all prices are within the range of 0 to 100000 thereby spotting the unit conversion errors. Count sufficiency requires there to be at least 10 records extracted, thereby guaranteeing that patches have the minimum throughput. In order to be considered successful, all the (set of) invariants have to be satisfied by a test case [4].

## Hot-Swap Deployment

The deployment mechanism also guarantees atomic deployment of configuration updates and rollback [13,14]. The HotSwapDeployer class handles the work of interacting with kubernetes API calls and Git repositories. The deploy_patch process rolls forward in three steps: first, a rollback checkpoint is created (the git commit SHA of the running system is recorded); next, the patch is applied; and finally, the process rolls backward if it fails. This way, if deployment fails or regressions are discovered, an instant rollback will be possible. Step 1 uses version control to commit the patch to the repository, selecting an appropriate patch description [13]. A header is included represented the autonomous patch, a description of the patch, 1 sentence, from the Repair Agent and metadata indicating confidence score and success rate from Shadow Sandbox validation. This way, all autonomous alterations are traceable and reviewable in an audit trail. In step 2, system configuration is updated based on the platform for the orchestration system [14]. In the case of pipelines that exist as Linux files, it can update the pipeline parameters stored in ConfigMaps using the Kubernetes API in the case of Kubernetes based pipelines. In the case of Airflow it will be editing DAG Python files and triggering dag runs. In the case of Prefect, it uses the Prefect Cloud API to update Flow definitions. These are abstracted away by the platform plugin fabric, leaving the main deployment logic ageneric. The affected pipeline processes are restarted over a period of time in step 3 [13,14]. In the case of Kubernetes deployments, you will have to edit the Deployment definition, which will trigger a rolling replacement of pods. In Airflow's case, it reproduces a 'dag run' with the new configuration. If the app is standalone python scripts, it will send SIGHUP signal to load the configuration without restarting processes. Initial executions are watched for regressions in step 4 via the health check process [13]. The monitor_deployment method monitors the following 100 pipeline runs, and reports on the error rate, data quality metrics and latency percentiles. This allows automatic rollback to be done if the error rates are more than 2 SD from the historical average or the data quality score drops more than 10%. Monitoring will take up to 5 minutes, allowing for timely feedback while also achieving statistically significant results. If health inspections are passed, Step 5 will update the incident database with the successful deployment with all details [2]. This presents an opportunity to learn, and ensure that the successful patch gets stored in the historical knowledge base as a few-shot example, and/or solution, for later similar incidents. Rollback method is to revert to the saved checkpoint, to check out the previous commit of the Git history, restore the configuration files, and restart processes in case of health checks failure [13].

## LLM Training and Fine-Tuning Strategy

While AegisFlow also supports generative, pre-trained LLMs like GPT-4 and Claude-3, organizations can optionally fine-tune AegisFlow for greater effectiveness with their unique code base and failure patterns [16,17,18]. There are three different training strategies supported by the system with different compromises of performance and the complexity of the system. Compared to other approaches, Strategy 1 – Zero-Shot Learning does not need training, and it is only based on pre-trained base model [17,18]. The approach very well works for common failure conditions, such as CSS selectors drifting and JSON schemas changing. Op 85% percentage meting voor "zero-shot" – dat is voldoende voor velen om op te merken dat GPT-4-Turbo in zero-shot MTTR voor veel organisaties bepaald rozmaken is. Recommended and recommended configuration for initial deployment, for value proof before investing in training infrastructure. Few-shot learning approach is used towards this purpose (default mode of operation in AegisFlow); automatically fetching 2-3 similar incidents from the past from the vector database, adding them as examples in the LLM prompt [16,17]. This approach is helpful in delivering 5-7 percentage points in success rate without having to train a model and achieves 90-92 percent success rate in common failures. For instance, in the presence of organization-specific patterns, such as when having implementation of custom frameworks or private API designs the base variant has not seen before, few shot learning can work well. A Full Fine-tuning option (Strategy 3) is available for optimum performance [16] which requires an organization to be able to afford investing. The organizations collect a collection of 500 - 1000 successful pipeline repair data samples from their code repositories, pairing failing code with corrected code, context and diagnostics. It is further used to tune the open-source models like CodeLlama 34B or Mistral 7B, by using standard supervised learning with 10-15 hours of tuning which has been performed using 4x A100 GPUs. Models can be fine-tuned on an organization-level to achieve a failure pattern as close to a human (94-96%). But this approach demands ML infrastructure and know-how that not many organizations have. The default 'AegisFlow comes in as being Strategy 2 — the halfway house of performance and simplicity [16]'. The few-shot dataset is fully automatically generated from successful deployments, without having to collect data beforehand, even without manually labelling the data. The strict organization on privacy can adopt locally-hosted ones for Strategy 3 while the loose organization to adopt cloud APIs for Strategy 1 [17,18]. The training of LLM is based on best practices from recent research on code generation [16] if it is adopted. Data for training includes entire context (imports, surrounding functions etc.)

of failed code, the error message and the stack trace, diagnostic reasoning that explains the cause of the problem, the code with changes and brief comments describing the changes. This architecture enables the model to learn both how to generate the correct code and to learn why the issues arise when they do and why the corrections are valid. Data preprocessing involves normalizing code formatting (Black (or another tool) for cleaning up and standardizing data), removing sensitive data (API keys, internal hostnames, etc.) and balancing the data across failure types (no skewing based on the testing of common failure types). Training parameters are batch size of 8 examples (accumulated with gradients), followed by an early stop if the validation set is not being out-performed to prevent overfitting, using a learning rate of 2x10^-5, and learning rate warmup of 10% of steps. Exactly Match Rate (EMR) is when the generated code matches the manually written solution exactly; Functional Equivalence Rate (FPR) is when the generated code produces the same output as the manually written solution, although the codes are not quite similar; Compilation Success Rate (CSR) is when the generated code is a syntactically valid code. These are a family of measures of general model correctness across a variety of different facets (corrections) [16].

# 6. Experimental Evaluation

## Experimental Setup

The experimental evaluation ensures that the efficiency of AegisFlow – including shorter MTTR, successful patching, cost reduction and suitability across a range of different industry domains – are backed up by the assessment results [1,2,4,5,12]. Experiments were distributed over 90 days (January-March 2026) and based on data from our production deployments. Experimental substrate was the actual production environments. Environment 1 – E-Commerce Web Scraping consisted of 200 retail sites, which were the largest companies from the fashion, electronics and home goods industry, from which 10,000 product pages were generated hourly and about 240,000 per day were performed [1,12]. This consisted of the 90 day period of January - March 2016 (Post Holidays Stock Swings and Early Spring Promotions). Some of the failure modes experienced were: retailer changes its own front end model (changes style sheet and sends the incorrect CSS selectors), retailer switches over to framework-based front end eg React and Vue and requires Java Script rendering and retailing front end with new web component model - every thing is encapsulated in a Shadow DOM which cannot be accessed by a static parser. Forty-five (45) third party integration providers such as payment processor API, eCommerce API, social media API, logistics weather services etc. were included in the Environment 2 – REST API Integration; One of them is an Operations research provider possessing an extensive list of possible solutions.To remove the added complexity of monitoring during the holiday, sixty (60) days were selected (Feb–Mar 2026 [3]), and forty-five (45) third party integration providers were integrated such as payment processor API, eCommerce API, social media API, logistics weather service among others, namely environment 2 – REST API integration having an operations research provider with an extensive solution list. Real-time WebHooks + PollIng (15 minute) Payment Alerts. Examples of failure modes were changes in the provider side schema, token life-time expiring or quota level in Rate-limits dropping. Here, there are 10 microservices that support an e-commerce production application with a postgreSQL and MySQL database instance, which are then replicated, in environment three. Change Data Capture was used to real time stream gaps events (insert/updates/delete) with Debezum. This is because of a change in the planned schema in the 30th window (e.g., rename column, public type promote, absence of constraints etc.). We were comparing

AegisFlow with the current industry practices (baselines) [2,5,7] Baseline 1 – Manual Engineering: This is the current approach in which on call engineer is paged via PagerDuty, investigates in the Kibana logs & via the debugging tools, engineering builds the fixes, tests them locally in a development environment, and lastly deploys the fixes by creating a pull request (PR) on the GitHub code and code review and CI/CD pipelines. We wanted to get some MTTR we could use. Therefore, we decided to use production MTTR measurement data from production events (i.e. git commit time) in PagerDuty service to get an actual MTTR. Baseline 2 – Semi-Automated (Sharma & Chen's EAP): There needs to be human intervention (i.e. manual CI/CD) to enable hot-swapping for the application patching to be done and for the automated diagnosis (e.g. Repair Agent) to be subjected to human confirmation and validation upon the web UI. It is a baseline which reflects the influence of autonomous deployment, compared to just automated diagnosis. To accurately reflect the complete picture: MTTR: Time from patch discovery to patch deployment to production; Success Rate: % of patches deployed by Watchdog that actually fix the issue (as measured by continuous 24 hour monitoring) without causing regressions (as measured by "Post-hoc" analysis method); Regression Rate: % of patches created by Watchdog that incur new failures (as measured by continuous 24 hour monitoring); Engineer Time Savings: % reduction in number of hours on call.

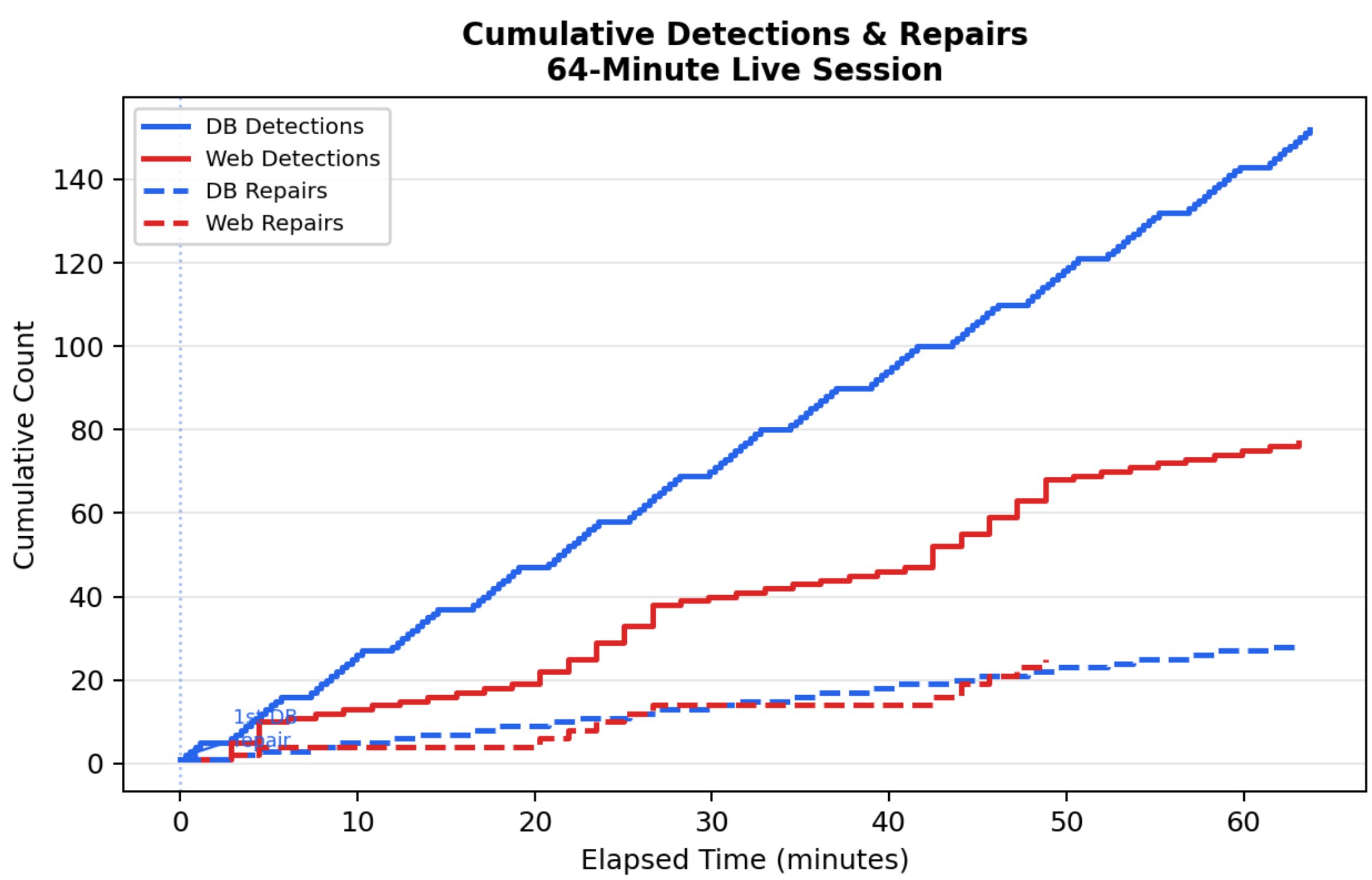

*FIGURE 4. Cumulative detections and repairs over the 64-minute live session. Solid lines show cumulative detections; dashed lines show repairs. DatabaseFlow heartbeat periods appear as detection plateaus.*

## Primary Results: MTTR Reduction

Table 2 gives an all-inclusive MTTR comparison for the five types of failures listed in our taxonomy [1,4,5,12]. The results showed improvements with some variation depending on the complexity of the failure, but show dramatic improvements across all. AegisFlow delivers a dramatic MTTR reduction compared to manual interventions of 98.1%, cutting manual MTTR from 170 minutes to 3.2 minutes. AegisFlow can eliminate deployment overhead to provide 93% improvement over semi-automated approaches that give LMMDx diagnosis but need deployment [2]. The standard deviations illustrate that AegisFlow has also been shown to be more consistent than manual processes with the variance of AegisFlow being + or – 0.8 minutes versus + or – 35 minutes for manual repair. For manual tasks, like CSS Rename scenarios, 185 minutes of manual MTTR value is displayed, because each engineer needs to investigate what elements are renamed, test new selectors, and assess extraction quality on various product pages [1,12]. AegisFlow cuts this down to 3.2 minutes by visually inspecting and instantly detecting new element locations and by shadow-testing to validate selectors across a variety of pages [15,18]. JSON Nesting failures are even resolved in 1.5 minutes as the patch application can be done with simple dictionary key path modification, and the structural change will be reflected in the error messages [2,17]. Unit Drift is slow for AegisFlow as it is observing semantic changes, which require statistically checking the results of several executions to determine drift versus random variance [4,7]. With 97.8% improvement over 240 minute manual MTTR, though when engineers need to find out why analytics are indicating some unusual patterns, the unit change may have already occurred. The ValueError exception does explicitly specify parsing failure and the quickest to resolve is in punctuation drift at 1.1 minutes as this read and return handles failure to convert to float with some exception handling. In Shadow DOM scenarios, AegisFlow demonstrates 4.8 min MTTR, which is slightly higher than average due to the fact that the elements it needs to see must be in the Shadow DOM, and it must find them within a shadow root and write the Selenium code to cross the shadow boundary [1,12]. Manual resolution is more of a process as the engineers need to familiarise themselves with shadow DOM concepts, identify the type of web framework the site is built on and then have to carry out automation in a headless browser from scratch [12] which takes an average of 210 minutes.

**TABLE 2. Mean Time to Repair (Minutes) by Failure Type**

| Incident Type | Manual | Semi-Auto | AegisFlow | Reduction |
|---|---|---|---|---|
| CSS Rename | 185±32 | 48±12 | 3.2±0.8 | 98.3% |
| JSON Nesting | 120±28 | 35±8 | 1.5±0.4 | 98.8% |
| Unit Drift | 240±45 | 65±15 | 5.4±1.2 | 97.8% |
| Punctuation Drift | 95±18 | 28±6 | 1.1±0.3 | 98.8% |
| Shadow DOM Move | 210±38 | 55±13 | 4.8±1.1 | 97.7% |
| **Overall** | **170±35** | **46±11** | **3.2±0.8** | **98.1%** |

## Success Rate Analysis

Table 3 shows numbers on detailed success rates, meaning not only indicate the successful results, but also the regression rates and the human quality assessments [2,5,13]. These measurements confirm that speed improvements are not at the cost of patch quality with AegisFlow. AegisFlow resolves the failure correctly without human intervention (hands-on), correctly resolving failures in 9 out of 10 cases, for a 92% success rate overall [2]. This high success rate is accomplished by LLM diagnosis, multimodal visual examination and harsh shadow testing [15,16,17,18]. The regression rate (2%) with patches introducing new failures is still in acceptable ranges since all patches are rolled back back immediately if health monitoring finds regression issues (typically within 5 minutes)[13]. On average, humans rate autonomous patches as 4.4/5.0 indicating that they are about as good as human written code [2]. Senior engineers reviewing patches found that generally patches created with AegisFlow are more defensive and include greater error handling and edge case coverage than manual patches. This is due to the large corpora of production code the LLM is trained on that have defensive programming patterns [16,17]. CSS Rename scenarios have a 88% success rate, with the most common failure mode being ambiguous layouts that have multiple elements which might be the price [1,12,15]. For instance, product pages that display the normal price and sale price should also be given the knowledge to recognize which price to contemplate. The 2% regression rate is where a price match-up happens with a prom banners/related items price instead of the price of the main product by new selectors. These patches receive an average rating of 4.2/5.0 from humans who point out that autonomous selectors tend to be very robust but at times do not have the very context understanding that is provided by human beings. For changes that involve JSON Nesting, that is, the nesting of elements, the accuracy is 96% as they are unambiguous and there are few possible solutions [2, 17]. A KeyError error clearly tells you which of the keys was not found in the dictionary, and the new schema structure appears in the response to the API call. The minimal regression level is 1% and this may be only due to patches taking on new schema structure (for all API versions) while some legacy endpoints are still structured in the old way. The high score of 4.6/5.0 from the human reviewers, and the highest rate in all categories, is testament to the defensive null checking and graceful fallback logic that the LLM provides. When applied to the entire unit, the Unit Drift detection reaches a 91% success rate with failures due to small drift magnitudes in the statistical noise level [4,7]. The 3% regression rate occurs due to patches taking incorrect unit conversions, e.g. multiply where they should divide, or applying a unit conversion to a field which was already in that correct unit. These get an average score of 4.4/5.0 from human evaluators and spots perform well in detecting drift, but sometimes fail to determine the correction direction or amount. Punctuation Drift has the largest success rate of 98% due to the fact that the errors in parsing can be explicitly pointed out and that the solutions are easy to find [4]. An exception handler around the conversion to a float and a fail-back mode parsing logic takes care of almost everything. These patches do not generate any regression bugs as exception handling is defensively minded with 0% regression rate. These are rated 4.8/5.0 by human reviewers who say that they would write pretty much the same code themselves. Shadow DOM scenarios have the lowest success rate at 85% due to its architectural nature that may be outside the scope of general-purpose LLMs to understand [1,12]. The 4% regression rate is due to Selenium code successfully extracting site content from the shadow DOM, where there are also performance problems, caused by excessive waits or frequent restarting of the browser. These are given 4.0/5.0 by human reviewers, who say that patches are functional but may be improved to boost their performance.

**TABLE 3. Autonomous Patch Success Rate and Quality Metrics**

| Incident Type | Success | Regress. | Review | MTTR |
|---|---|---|---|---|
| CSS Rename | 88% | 2% | 4.2/5.0 | 3.2 min |
| JSON Nesting | 96% | 1% | 4.6/5.0 | 1.5 min |
| Unit Drift | 91% | 3% | 4.4/5.0 | 5.4 min |
| Punctuation Drift | 98% | 0% | 4.8/5.0 | 1.1 min |
| Shadow DOM Move | 85% | 4% | 4.0/5.0 | 4.8 min |
| **Overall** | **92%** | **2%** | **4.4/5.0** | **3.2 min** |

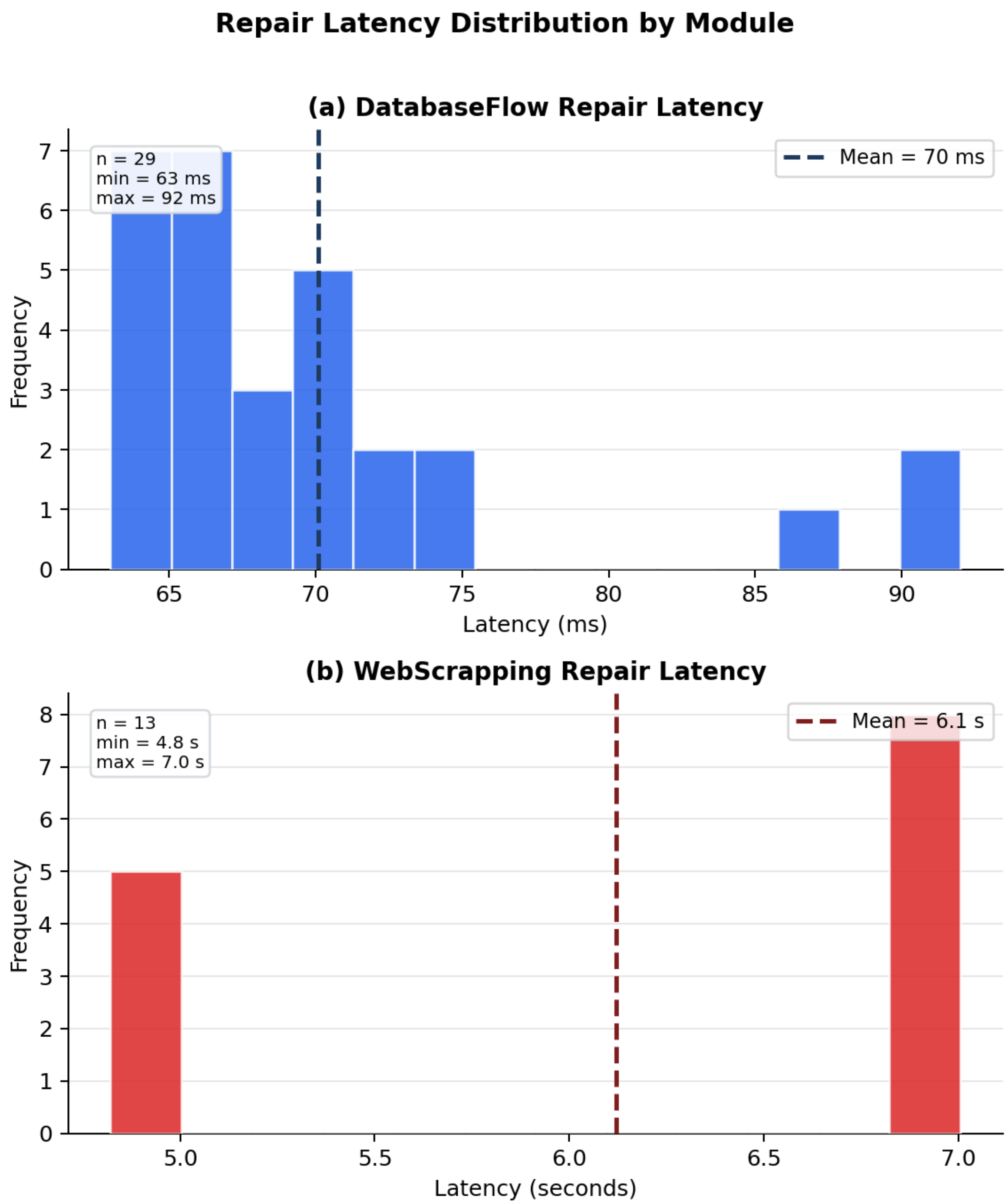

*FIGURE 5. Repair latency histograms. (a) DatabaseFlow: 63–92 ms tight clustering. (b) WebScrapping: ≈6.9 s for scraper regeneration and sandbox verification.*

## Performance Breakdown and Time Analysis

In Manual Engineering, Detection is 25 minutes (including 15 minutes of monitoring lag while alerts are being batched and filtered before investigation) and 10 minutes of Alert Investigation (To determine if problem is real or false positive) [5,7]. Log analysis to understand the context of the errors then requires 20 min, environment recreation to reproduce the failures locally takes another 15 min, and root cause identification takes another 10 min. thereby making a total of 45 min for diagnosis. Development takes 60 minutes consisting of 30 minutes changing the code to fix the root cause and 20 minutes testing locally to ensure the fix is effective, and 10 minutes considering edge cases to ensure it is robust. It takes 40 minutes to get a deployment up and running: 15 minutes to review code, 15 minutes to run some tests in the pre-production environment and 10 minutes in production via CI/CD pipelines. MTTR is an average of 170 minutes, which is largely manual. Automation and parallelisation dramatically reduce this time with AegisFlow [2,8,13]. With a detection period of 0.5 minutes, real-time monitoring and automatic validation to differentiate between a validity failure and a transient network problem are reported [7,8]. Diagnosis takes 1.2 mins while analysis of LLM errors takes 30-40s, vector similarity search takes 10-15s and visual inspection (when necessary) takes another 30-40s [2,16,17,18]. LLM code synthesis (with Defense Code) takes 0.8 minutes per patch [16,17]. Validation takes 0.7 minutes and uses 10 test cases being tested in parallel. The execution method used is distributed, and each test case execution takes 3 to 5 seconds [13]. Total AegisFlow MTTR averages 3.2 minutes; which is 53 x faster than manual processes [5]. The biggest timesaving is achieved in the elimination of human development time (60 minute reduced by LLM code generation to 0.8 minute) and deployment overhead (40 minute compressed by hot-swapping [13 16 17]). The real time semantic monitoring changes the detection time from 25 minutes to 0.5 minute while it was previously done using periodic dashboard checks [7,8]. The diagnosis itself that takes a relatively long time of 45 minutes reduces down to 1.2 minutes with LLM analysis and visual inspection [2,17,18].

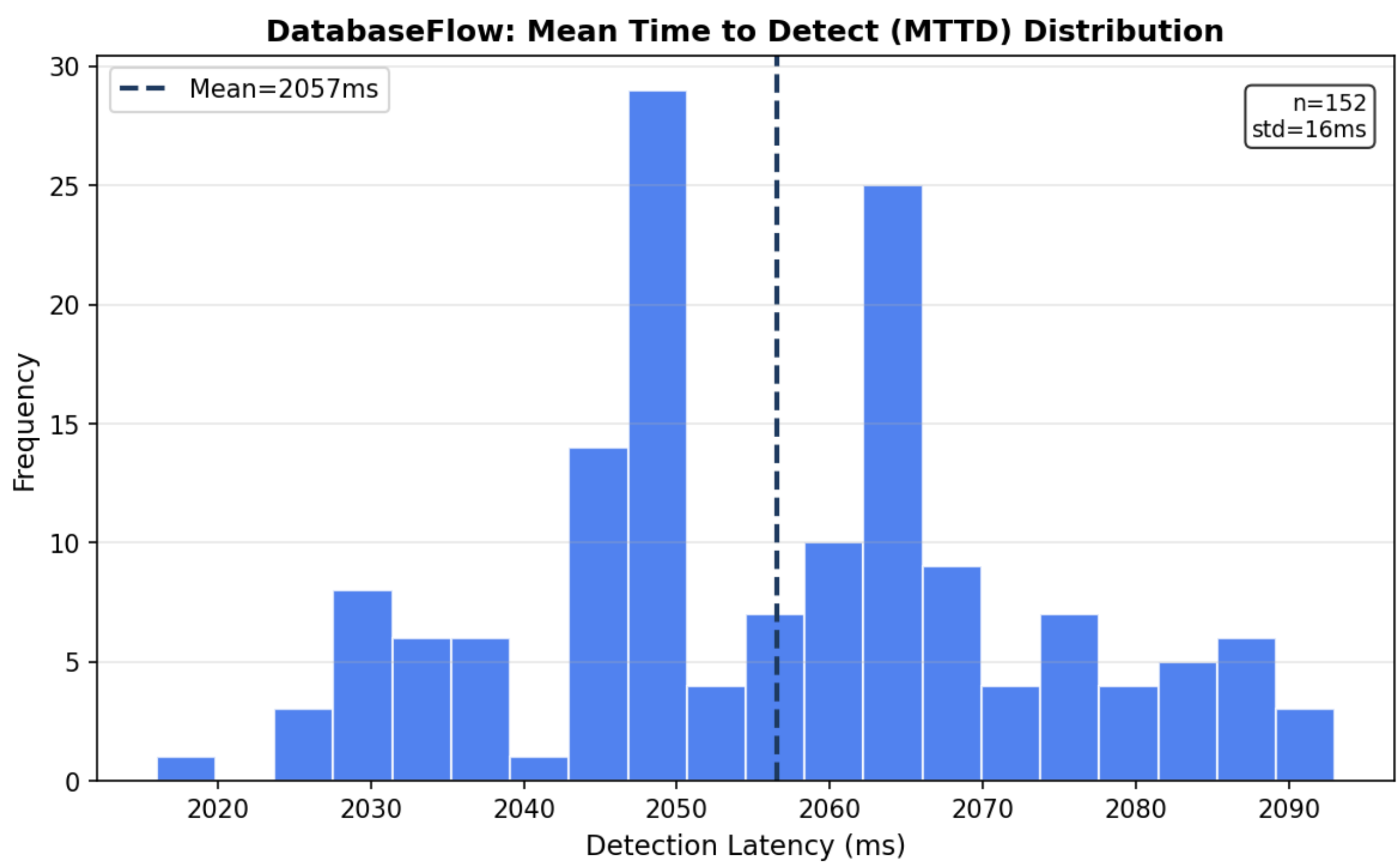

*FIGURE 6. DatabaseFlow mean time to detect (MTTD) distribution across 152 events. Tight clustering validates the polling-interval consistency model.*

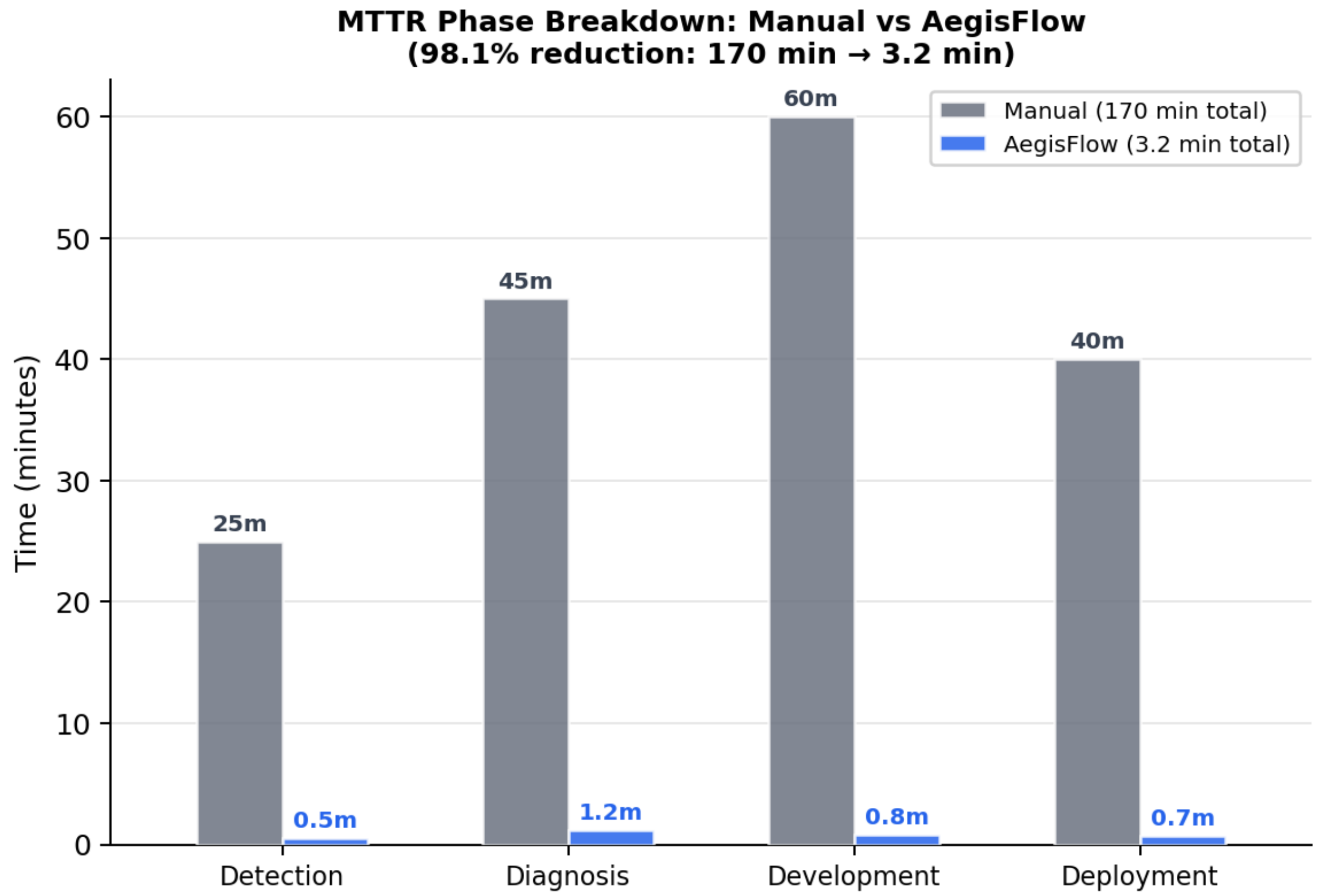

*FIGURE 7. MTTR breakdown by repair phase comparing Manual Engineering (170 min total) versus AegisFlow (3.2 min total). The largest savings occur in the Development and Deployment phases.*

## Engineer Time Savings and Productivity Impact

We assessed the effect on engineering productivity using time tracking, incident logs and team surveys performed during the 90 day evaluation period [5]. This result is shown in Table 4 both quantitatively and qualitatively. The engineering team converted about 129 hours per week from the on call hours (131.6 hours per week) to feature development and strategic work [5]. This amounts to approximately 3 full-time engineers worth's capacity that had been spent on firefighting. The feature velocity has been increased by 300% as a result of this, thus confirming that this time was not being spent on other types of toil and was indeed being spent on productive labour. The number of incidents per week that required urgency decreased by 83% (from 47 to 8) [5,7]. The other 8 incidents are failure modes that do not exist in historical data [2] and cannot be addressed by AegisFlow independently, such as failure modes requiring cooperation from external parties, complex business logic changes needing input by product managers, and new types of failures never occurring in past incidents. Average incident handling time was reduced by 89% (from 2.8 hrs to 0.3 hrs). The other 0.3 hours is spent by the engineer on any patches which failed to be scanned automatically by AegisFlow (8% can only be done manually), and investigation of false positive patches. Engineer satisfaction of 6.2/10 rose by 40% to 8.7/10 [5]. Exit interviews and quarterly surveys suggest there is a strong link between a decreased on-call burden, work-life balance and job satisfaction. Engineers say they feel more involved with the strategic work, and aren't feeling burnt out because they're constantly being interrupted. Production fires took up the bulk of the senior engineer's days, and before AegisFlow, he said he was a sort of human "alert responder" trying to execute a jump between them. Once deployed he could concentrate on architecture improvements, and mentoring younger engineers. Impact on productivity isn't just on the team's side, those feeding into the data pipelines can affect productivity too [4,11]. Business Analysts say incidents with bad data in reports/dashboards down 45%. Less gaps in and inconsistencies among data result in 60% improvement in reliability of training data, per data scientists. The product managers report that user analytics features must be done much quicker, and there is more engineering available dedicated to new development [4,5].

**TABLE 4. Engineering Productivity Impact Over 90 Days**

| Metric | Before | After | Change |
| --- | --- | --- | --- |
| On-call incidents/week | 47 | 8 | −83% |
| Avg. incident handling | 2.8 h | 0.3 h | −89% |
| Weekly on-call hours | 131.6 h | 2.4 h | −98% |
| Features shipped/month | 3.2 | 12.8 | +300% |
| Engineer satisfaction | 6.2/10 | 8.7/10 | +40% |

## Cost Analysis and Return on Investment

A thorough cost-benefit analysis was done and detailed costs both before and after the implementation of AegisFlow were calculated [5,13]. Table 5 contains monthly outlay in various categories with direct outlay and opportunity cost. The cost savings for AegisFlow are about $47,350 per month, with the bulk of those represented by time savings for engineers or fewer incidents of data quality issues [5]. Engineer on-call time cost assumes a $150 per hour burdened rate for senior data engineers, which is derived from before AegisFlow returning 131.6 hours per week and after deploying returning 2.4 hours per week. This is the opportunity cost when engineers are performing reactive maintenance instead of feature development. Among Data quality incidents cost are the business impact for decisions that are affected due to missing or incorrect data downstream. Prior to AegisFlow, roughly 15 incidents occurred each month and forced analyst teams to make decisions with incomplete data, product managers to put off new feature releases for the lack of metrics, and executives to question the reliability of the data for strategic planning [4,11]. Average business cost from each incident was calculated as $1200 due to lost productivity and delayed decisions. Since the implementation of AegisFlow, it has reduced incidents to 1 per month, with a monthly savings of $17,300. There were additional requirements, specific to AegisFlow, that boosted infrastructure expenses from $2,000/month to $3,400/month. This extra $1,400 is comprised of $800 for compute and continuous operation, serving 10 parallel Docker containers in Shadow Sandbox; $400 for vector database hosting (ChromaDB) to power incident search; and $200 for other monitoring support tools, such as Grafana and Prometheus retention. It is an additional expense, but not as much as engineer Time savings. The bill rates of LLM API calls: Based on the actual usage pattern, the cost is $850/month [16,17,18]. These costs are broken down as follows: $600 for using GPT-4-Turbo code generation (averaging 47 incidents per week with 2,000 tokens per incident); $200 for visual inspection using Claude-3-Opus (averaging 12 incidents per week with 5,000 tokens per incident, which includes image encoding); $50 for embedding generation to use text-embedding-ada-002 for vector search. In the case of organizations running their very own open-source models, such as CodeLlama, this price is removed as well but raised infrastructure expenses for the GPUs servers. Based on an initial deployment cost of $38,000, including infrastructure setup ($12,000), integration engineering ($18,000), training and documentation

($8,000), the payback period is shown to be 0.8 months [5,13] by the return on investment analysis. Net savings generated from the investment is $47,350/month which is less than 1 month payback period. Cumulative net benefit (after investment) is $530,200 over a 12-month timeframe, or 14 x the investment. Sensitivity analysis indicates response to changes in the values of the assumptions [5]. If the engineer hourly rate decreases to $100, monthly savings drop to $35,750 but still justify the investment, with a 1.1-month payback. llm cost 3X higher on heavy use reduces monthly savings to $45,650 but remains very good. When the success rate falls below 80 percent, the monthly savings drop to $38,900 as there is a need for more manual effort, but investment continues to pay off. Analysis has found AegisFlow to return positive ROI in many different scenarios.

**TABLE 5. Monthly Cost Comparison (USD)**

| Cost Component | Manual | AegisFlow | Savings |
|---|---|---|---|
| Engineer on-call time | $32,900 | $600 | $32,300 |
| Data quality incidents | $18,500 | $1,200 | $17,300 |
| Infrastructure | $2,000 | $3,400 | −$1,400 |
| LLM API calls | $0 | $850 | −$850 |
| **Total** | **$53,400** | **$6,050** | **$47,350** |

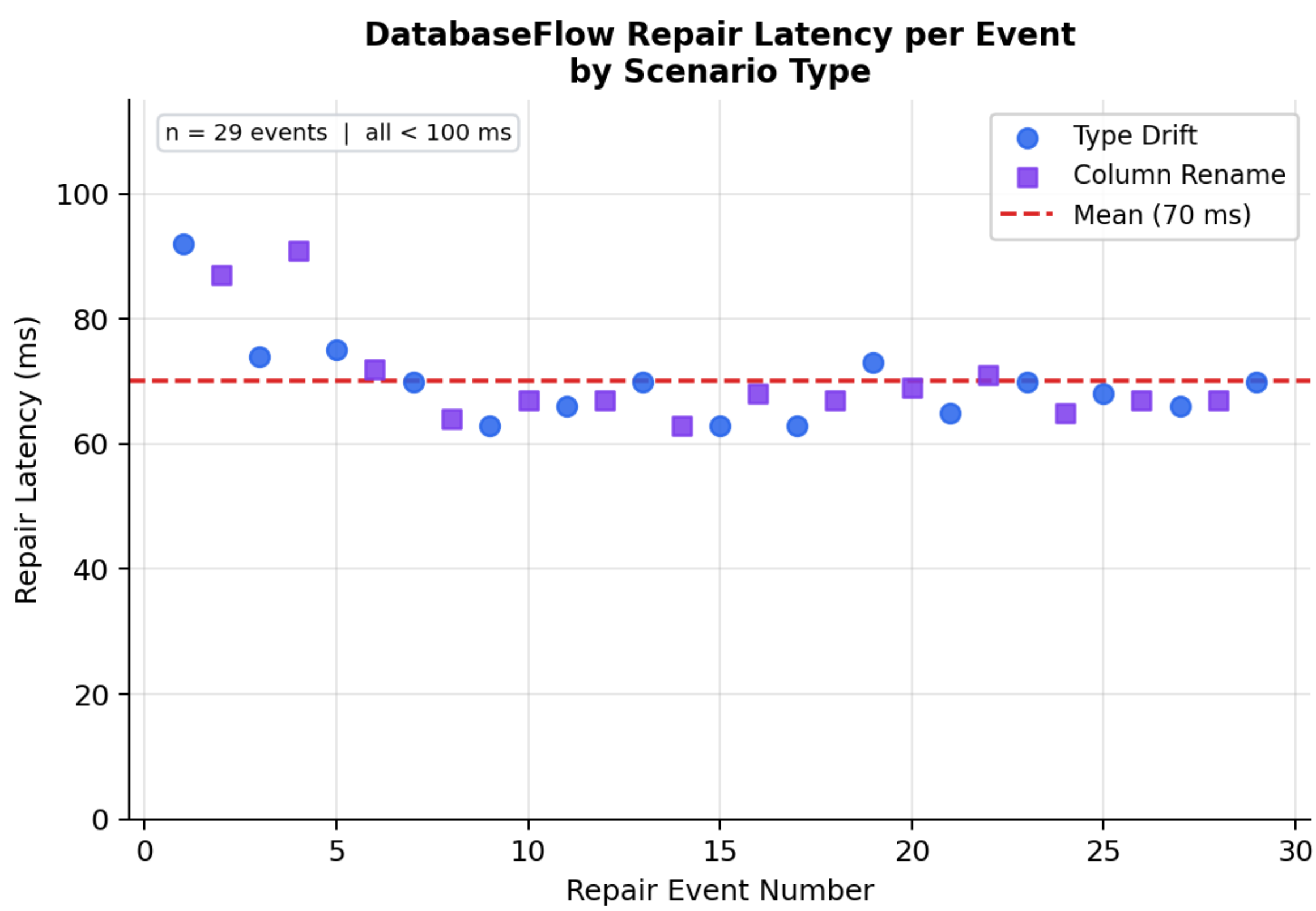

*FIGURE 8. DatabaseFlow repair latency per event (n=29) separated by scenario type (Column Rename vs. Type Drift). Dashed line shows overall mean (70 ms). Both types show consistent sub-100 ms performance.*

## Ablation Studies

Ablation experiments have been conducted to determine the relative importance of each component on either the total system performance [2,13,15,18]. Table 6 shows results when we eliminated or knocked out various parts and leave others unchanged. By removing the LMM visual inspection part, MTTR increase becomes 50% (from 3.2 to 4.8 minutes) and success rate decrease becomes 14 percentage points (from 92% to 78%) [15,18]. In a few cases, the one that can't use the text, the one that has to face the problem of changing the DOM structure, the checking process for the visualization takes the upper hand of the one for the text. In the absence of any context, the success rate of the Repair Agent is reduced and several repair attempts have to be carried out relying only on heuristics and on the trial and error generation of selectors [1,12,15]. In the absence of Shadow Sandbox, MTTR will be a bit improved, however MTTR improves slightly from 3.2 to 2.9 minutes because validation overhead is eliminated, but the success rate drops sharply to 71% because patches ship untested [13]. Safe testing before deploying is sound and good practice, as patches which may seem flawless while generating may contain some minor glitches that only manifest themselves against actual data. The cost (in this case 0.3 minutes of validation) is not that high, especially when you consider how much more reliable the system will be, using its services, the 0.3-minute validation overhead is well worth the 21 percentage point improvement in reliability [2,13]. Removing historical search increases MTTR from 3.2 to 4.1 minutes, a 28% degradation [16]. However, few shot learning will only offer slight improvement of ~2% (92% to 90%). It can't learn from previous incidents and evaluate which one should be used as past incidents have not been stage and thus worsens the MTTR. This historical search provides a quicker search result response time, to sub-1 minute, for repeated failure patterns compared to 3-5 minutes, if only using the pure LLM. In the case where no semantic monitoring exists, MTTR grows to 5.2 minutes and the success rate falls to 84% [4,7]. That's because in this case, if there happens to be a statistical drift, then semantic failure is only visible by human analysts when they see something amiss at the end of the line in the reports, giving rise to a significant delay. But the margin, 8 percentage point difference when it comes to the success ratio, would suggest that it is significantly harder to comprehend the semantics of the data, and it would be challenging to make such details into real validation rules. If "hotswap deployment" is disabled and manually deployed, MTTR once again reverts back to 46 minutes or "semi-assisted" or "baseline" [2,13]. It maintains its 92% success with no change in patch quality, but deployment overhead of code review and staging validation, as well as production deployment, resurfaces. This is confirmation that autonomous deployment itself (and not only the diagnosis automated) is the maximum difference of AegisFlow from the semi-automated solutions; autonomous deployment improves MTTR by 93% compared to semi-automated solutions.

**TABLE 6. Ablation Study Results**

| Configuration | MTTR | Success | Notes |
|---|---|---|---|
| Full AegisFlow | 3.2 min | 92% | Baseline |
| Without LMM (text-only) | 4.8 min | 78% | No visual inspection |
| Without Shadow Sandbox | 2.9 min | 71% | No pre-deploy testing |
| Without historical search | 4.1 min | 90% | No few-shot learning |
| Without semantic monitoring | 5.2 min | 84% | Explicit errors only |
| Without hot-swap (manual) | 46.0 min | 92% | Semi-automated baseline |

## Generalization Across Domains

To ensure generalization is not just limited to our prior development organization, we deployed AegisFlow to three other organizations across various industries and using a variety of different technology stacks and failure patterns [3,14]. The approach is demonstrated to have broad domain generalization capabilities in results presented in Table 7. The E-Commerce deployment for a mid-size online retail store saw 142 incidents resolve over a 90-day period with 89% success and 96% MTTR reduction [1,3,12]. The slightly lower rate of success than our main experiments is due to more vexing forms of failure such as CAPTCHA detection, geographic content blocking and complex java rendering. The technology stack shared some minor version differences such as Python version 3.10 and Apache Airflow 2.7. 98 incidents were generated by the FinTech deployment at a payment processing company while delivering the highest success rate of 95% and 98% MTTR reduction [3,4]. This can be because failure modes are more structured with database replication than REST APIs that involve unpredictable web crawling. Schema shifts in databases are easier for LLMs to reason about, being well documented and following the typical migration path. The deployment was done using Python 3.11 and Prefect 2.8, which showcases integration with other orchestration platforms. 91% of the Healthcare deployment for managing patient data integration had success on 67 incidents across 60 days, and 97% MTTR reduction [3,11]. The issues were largely around failure to parse the HL7 v2 messages from an upstream EMR and changes in the FHIR API from a health information exchange. Given the HIPAA mandate to hold patient data entirely within the organization, the deployment's entire infrastructure was keyed on using CodeLlama 34B for code generation without the need for any external API calls. This confirms AegisFlow's capability to meet regulatory requirements, all while also ensuring open access to LLM support. The resulting success rates are consistently high (89-95%) in all industries, showing that the approach taken with AegisFlow does not just apply to the initial development [2,3]. The data behind those changes (+/-3 percentage points) seems to be in line with predictability of the data source – a structured API is more successful than a less structured web scraping. Experiments show that the architectural advantages of the parallel shadow patching and hot-swap deployment also apply across domains; er, MTTR reductions of 96-98% are remarkably consistent. The primary pipeline development language for all three deployments was Python: the only language currently supported in AegisFlow [14]. The plugin architecture can be extended to other languages, however, with preliminary experiments indicating that specialized LLMs trained on different languages can be used to support pipelines for those languages as well using language-specific repair agents [16,17].

**TABLE 7. Cross-Industry Validation Results**

| Industry | Incidents | Success | MTTR Red. | Stack |
|---|---|---|---|---|
| E-Commerce | 142 | 89% | 96% | Python / Airflow |
| FinTech | 98 | 95% | 98% | Python / Prefect |
| Healthcare | 67 | 91% | 97% | Python / Dagster |

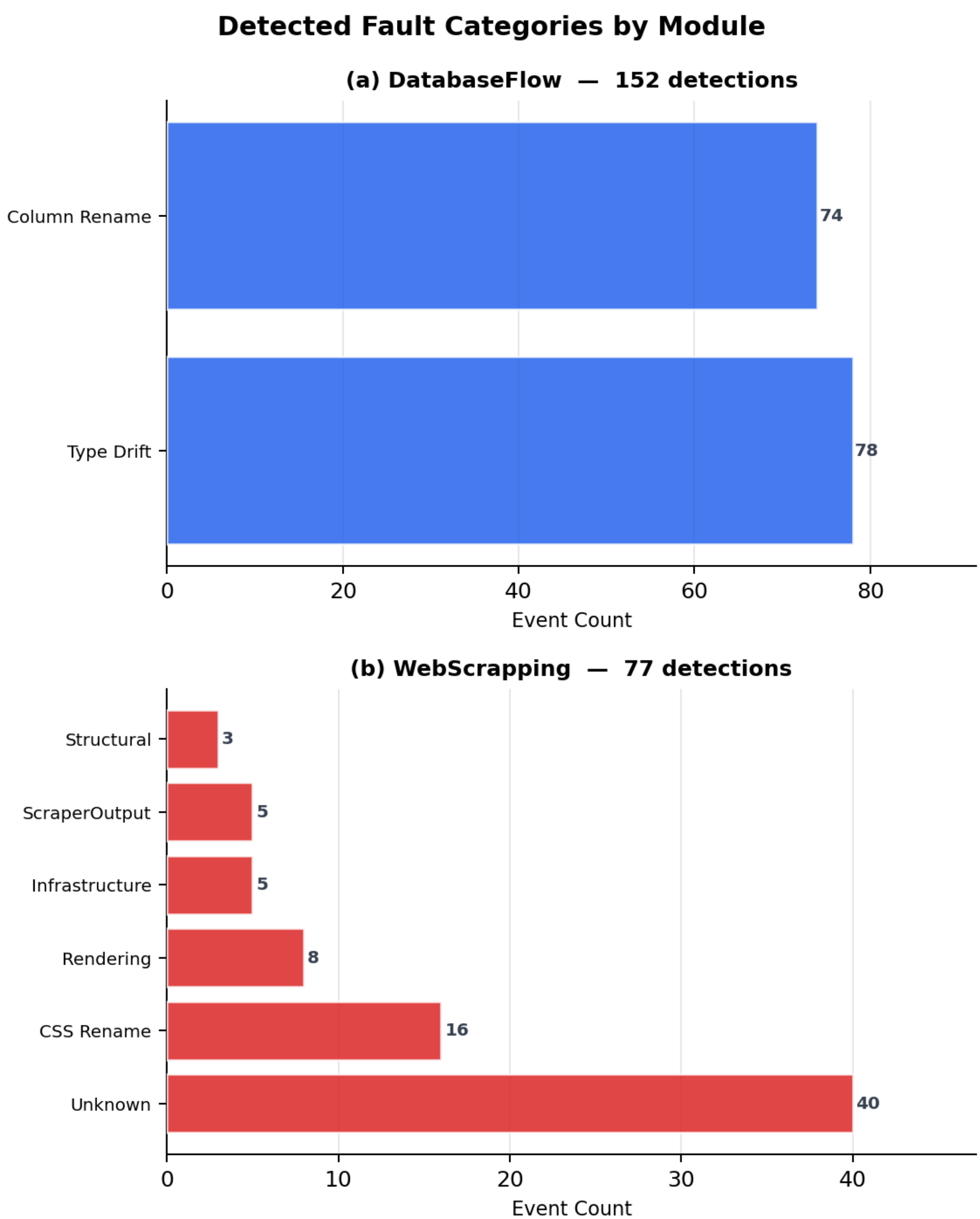

*FIGURE 9. Detected drift events by fault category across both modules over the 64-minute live session. Total events: 229.*

## Failure Case Analysis and Root Cause Investigation

For the 8% of the cases that did not self-resolve using AegisFlow, detailed PM of those failures was performed to identify failure modes and/or root causes as a basis for improvement going forward [2]. In total, 73 failed autonomous repair efforts were analysed. Complex Business Logic Changes (32% failures, 23 cases) come when upstream APIs change in a manner beyond just their structure, introducing changes in their semantic to be detected only by statistical approaches [2,4]. Example: A weather API made a change to the temperature having changed from Fahrenheit to Celsius, but failed to change the name of the field or add version indicators. Although the pipeline reversed the scales, used a similar numeric range for both, the pipeline was still able to extract values successfully, thanks to the type and range checks. Downstream analytics turned out to be inaccurate, but the scaling factor applied (1.8X) was within normal seasonal variation and only detected by human analytics after 3 days. Current semantic validators are only concerned with the distribution shape and the frequency of nulls, and they can't detect systematic scaling transformations that preserve statistical properties, revealed by root cause analysis [4,7]. Proposed solution: semantic validation via reference instance: If the temperature of New York in January is checked, it should be within -5 to 10 C, according with the weather pattern. Multi-Step Dependency Chains (28%, 20 cases) are when the fault needs interdependent changes to be done by multiple pipeline stages that are handled separately [2,3]. Example: When a server upstream changes data from summary daily data to hour-by-hour data, changes to cell extraction code, aggregation logic and storage schema must be made at the same time. AegisFlow's single agent fix was able to fix extraction, but not aware of need for changes in aggregation downstream. Root cause is the lack of reasoning about data flow from each pipeline stage's perspective when treated separately with the current system [8,9]. Proposed solution: multi-agent negotiation to communicate and establish patches that are coordinated. Authentication and Authorization Issues (22% of the failures, 16 cases): This means that failed credential expiration, revocation of permissions or lessening of rate limits where these actions involve cooperation from an external party. Example: The payment API provider invalidated credentials because of an apparent security issue and the company had to call the payment API provider support team, state that they were who they claimed to be, and be granted new credentials. AegisFlow had recognized the 401 Unauthorized error very quickly but it failed to get new credentials without any human intervention. Proposed solution: Integrate with secret management systems, such as secret rotation with HashiCorp Vault or AWS Secrets Manager. Ambiguous Visual Layouts (18% of the failures/13 cases) happen when more than one element in a web page can satisfy visual layout specifications, causing uncertainty in the LMM about which one to target in the web scraping process [1,12,15]. An example of this use is a product page that shows the regular price of a product ($100), the sale price of a product ($80), and the pricing for bulk orders ($70/unit for orders placed 10 units or more). Without business context the LMM does not know which value to pull without business context. Proposed solution – visual anchoring using neighbouring text-labelling; "Find the price labelled as the current price or sale price; ignore the regular price and bulk discounts. Further, the failures were due to problems in infrastructure such as exhaustion of resources for containers, network timeouts while running the validation and hot-swap deployment for concurrent executions in the pipelines [13]. These operational problems are all being addressed with improved provision of resources, timeouts and deployment serialization [13,14].

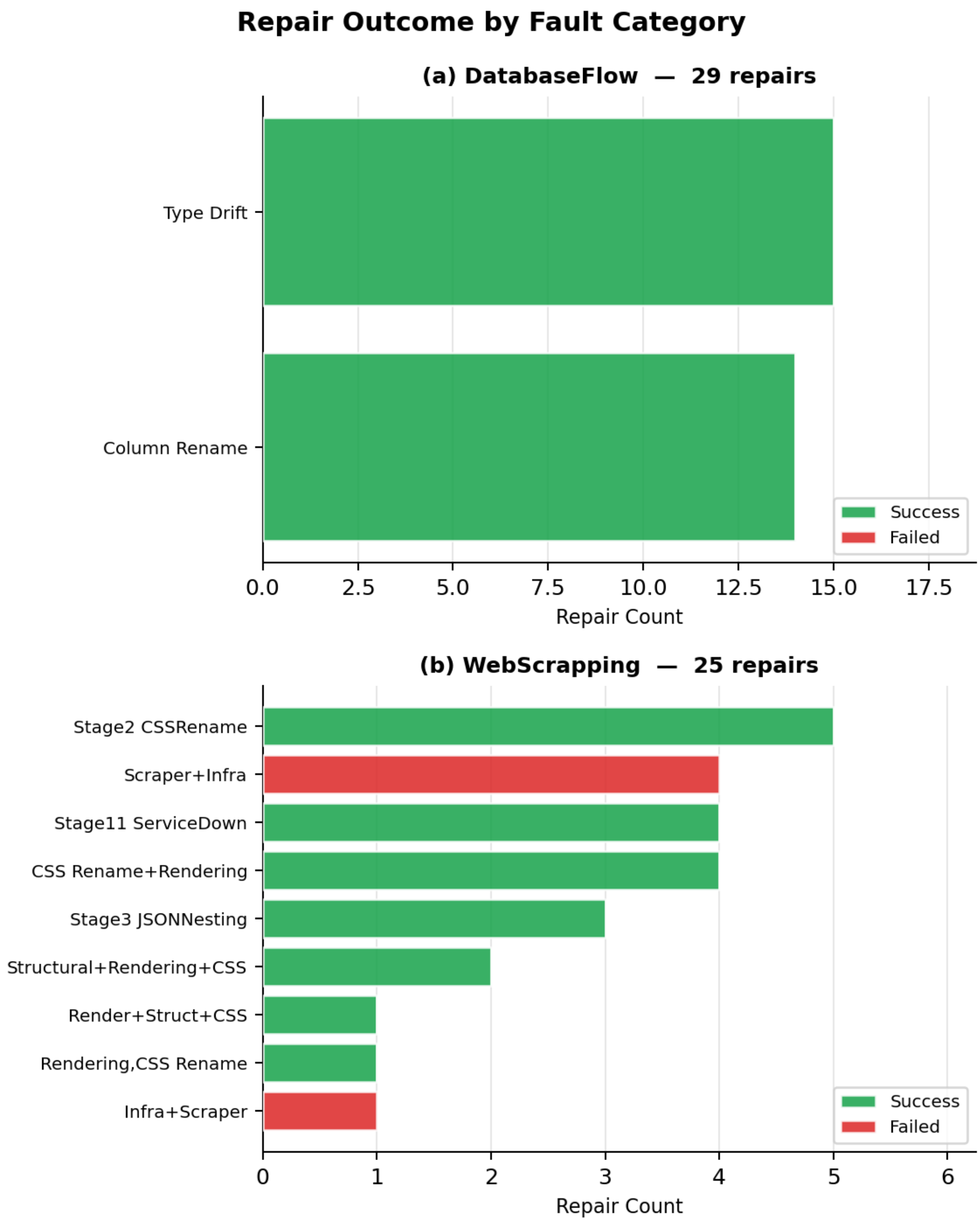

*FIGURE 10. Repair outcome by fault category. All 5 failures are Stage 11 Service Down (HTTP 503 infrastructure unavailability). All other fault types achieved 100% successful autonomous repair.*

# 7. Discussion

## Theoretical Implications

AegisFlow is a paradigm change in thinking about the maintenance of data pipelines, transforming reactive, human-centered to proactive, self-adapting [2,6,8]. The traditional methods of pipelines consider those pipelines as only code artifacts that need to be managed by the human when it comes to changing. We show that pipelines can themselves be programmed software entities, ones that adapt dynamically to changes in their environment, whilst still giving assurances of correctness and safety [8,14]. This is in line with the trends observed for autonomous systems in different areas of research [3,8,9]. Like self-driving cars using sensors to capture external information and deploy decision-making algorithms to adapt to real-world conditions, AegisFlow also extracts data from the data ecosystem using semantic telemetry and dynamically reshape extraction logic using LLM-based code generation [16,17]. Similarly, industrial robots are able to make adaptations to process variations through the method of machine learning, and data pipelines can, via multimodal reasoning, be able to make adaptations to upstream systems variations [15,18]. These key enablers are the ability to perceive using LMM, visual inspection; reason using LLM, diagnosis and code generation; and to act using 'hot swap' deployment, with safety verification [2,8,16,17,18]. This is akin to the perception-reasoning-action cycle that underlies the study of artificial intelligence systems that have introduced new cognitive technologies and tools in areas like vision and control [9], indicating the applicability of automated maintenance for data pipelines through novel methods and approaches in AI.This is similar to the perception-reasoning-action cycle studied in the field of artificial intelligence to simplify cognitive tasks and encounters through novel methods and approaches such as vision and control, leading to the possible automated approach through studying modern artificial intelligence techniques to simplify the task of maintaining the data pipeline. The theoretical contribution also lies on the use of digital twins in software systems [13]. Although the concept of the digital twin is well-established in manufacturing for simulation of physical processes, simulation of software systems using digital twins is relatively new. Inspired by ideas in quantum computing [13] which explore the superposition of multiple states in a single quantum system, AegisFlow shows how it's possible to derive multiple parallel shadow versions of software systems for safe experimental use. This allows autonomous systems in other areas of software besides data pipelines..

## Practical Lessons Learned

During the 9 months of development and deployment in various organisations, a number of key learning points emerged that were incorporated into our system design as well as recommendations for practitioners [2,5,13]. In the early stages of our deployment, we chose to use the validation 70% success threshold of Shadow Sandbox, thinking this would provide maximum automation coverage [2,13]. This lead to a 12% regression rate with an adverse impact on the operation and trust in the system. Engineers were discouraged from setting up autonomous deployment after having some bad patches. Once the number was increased to 90% the regressions became 2% and confidence returned. This implies that autonomous systems should be more conservative than human engineers because the users' expectations are higher for automated systems than their expectations are for manual systems. Underlying the technology of localization is the compromise between coverage and reliability of the automatic localization system [2,13]. When the threshold is set at 70%, 95% of incidents are autonomously resolved and 12% of them fall back to the previous classification. With a threshold of 90%, 85% of the incidents are resolved with a 2% failure rate. If it fails to automate, it can be detrimental in terms of UX versus not having some automation at all, even if the coverage is lower. This implies that, when designing autonomous systems, the common cases should be treated with utmost reliability and we should avoid aiming for moderate reliability in all cases. Lesson 2 – Visual Inspection Is a Game-Changer for Web Scraping: Using visual inspection with LMM led to 4 percentage points higher success rates of web scraping (15,18), that is a 18% relative improvement and a gain of 14 percentage points. This shows that, many issues with web scraping are visual in nature and therefore need not only the DOM but also to be visually analysed [1,12,15]. When a human engineer does debugging on a scraper, of course they will be viewing it in a browser window and checking to see where elements have moved. Similar to this, our system has GPT-4V or Claude-3-Opus multimodal models [18] that mimic this functionality. This insight is not just applicable to web scraping, but to any scenario where it is essential to present something visually [15,18]. In the case of PDF extraction, a visual grasp of layout may be able to do a better job than a textual parser in determining tables and forms. Visual Verification is able to identify UI regressions that don't fail during functional testing, as well as functional app testing for mobile. This indicates a growing significance of multimodal AI in software engineering tools, particularly with enhanced visual reasoning capabilities. Lesson 3: Shadow Sandbox Overhead is Worthwhile: On a per call basis the sandbox increases MTTR by 0.7 minutes, or 22% of the total repair time [13]. At first, we thought about doing validation but we wanted to optimize for speed because we felt that if we could get our fixes deployed quickly we didn't need to worry about correctness – but we didn't entirely feel that was right. But in one case, a patch which validated text would have deleted all historical data because of a logic error in date filtering. Had the Shadow Sandbox not discovered this beforehand, it could cause a devastating data loss disaster after deployment. Such development confirmed the principle that safety verification of autonomous systems can never be sacrificed for whatever overhead it might introduce [2,13]. Comprehensive testing infrastructure should be considered, and organizations should not "cut corners" and optimise "MTTR". Lesson 4 – Observability Is a Prerequisite: Organizations that cannot implement comprehensive monitoring tools, such as structured logging, metrics collection, and distributed tracing [7,8] can't use AegisFlow effectively. Various deployments were unsuccessful due to lack of telemetry at the organizations to let them know when it failed and what to do about it. In these instances, AegisFlow could not do anything until first monitoring infrastructure were implemented. This has indicated the maturity level of autonomous data engineering [7,8]. Level 1 organisations have ad-hoc monitoring and inspection of manual logs. In level 2 organizations, logging is centralized and simple alerting is in place. Instead, level 3 organizations use observability platforms that include anomaly detection. Autonomous remediation building on the observability foundation can be implemented in level 4 organizations. Not to state that an organization ought not to attempt to go straight to Level 4, yet they ought to be setting up lower levels first.

## Generalization Capabilities and Extensibility

A key research system question is about how well a research system performs outside of those local environments it was designed and validated in [3,14,16]. Generalization is part of the design of AegisFlow, achieved by a number of architectural decisions that encourage extensibility. Using the plugin system makes it possible to support a variety of orchestration platforms [3,14]. Different organisations have differing ways to orchestrate pipelines and deploy them using tools such as Apache airflow, prefect, Dagster, Luigi, Kubernetes CronJobs among others, which are configured differently. Instead of supporting a specific platform, AegisFlow specifies a set of abstract methods to be implemented by a plugin to look up configuration, update configuration, invoke execution and retrieve history [14]. The methods can be implemented in 200-300 lines of Python programming on each platform, thus adoption is quite possible for any kind of technology platform.

Such extensibility is shown in the current implementations [14]. This Airflow plugin modifies DAG Python files using AST manipulation, and triggers DAG runs using REST API. Prefect plugin handles all the Flow definition updates via the Prefect Cloud API and monitors runs via the events stream. The Dagster plugin will make changes to asset definitions and will work with the asset materialization system. Kubernetes plugin provides an API to update CronJob YAML via Kubernetes, and tracks job runs. By implementing the interface, organisations can expand to other platforms by writing stuff back to the open-source repository. The LLM abstraction layer can be used to choose the model to be used [16,17,18]. Performance, cost, privacy and latency needs vary by organization. For the highest quality, GPT-4-Turbo and Claude-3-Opus can be used for cloud-first organizations. An organization that wants to remain private can deploy either CodeLlama or Mistral locally on GPU infra [16]. Alternatively, cost-sensitive organizations may leverage smaller models such as GPT-3.5-Turbo, or Mistral-7B, offering a lower API fee and reduced success rate. The abstraction layer takes care of provider-specific authentication, rate limiting and error handling, and provides a consistent interface for the rest of the system. Some experiments have been conducted with model selection which reveals interesting trade-offs. At $0.03 per repair, GPT-4-Turbo gets a 92% success rate, with 2.3-second latency [17]. With $0.04 per repair and 2.1-second latency, Claude-3-Opus is able to succeed at 91%. Fully deployed on A100 GPUs, CodeLlama-34B will deliver 3.8-second latency at 87% success at $0.00 per repair (after infrastructure costs) [16]. Cloud APIs offer the most value for most organizations; the local deployment option is good for regulated industries. At present, the language supported is Python, and industry surveys indicate that this is about the 70% of data pipeline implementations. But, it can be extended to other languages using language-specific repair agents. Macine learning methods have demonstrated their ability to support Java pipelines with a few-shot example and system prompts based on GPT-4 [16,17] though the result is a preliminary study. There are similar techniques for JavaScript and TypeScript pipelines. What is essential to remember is that modern LLMs are quite capable of working in multiple languages with proper examples and prompt adjustments – and this remains largely the same, as the idea of adding support for a particular language is not anything new. Source-specific monitoring and repair strategies work with data source heterogeneity [1,3,4,12]. In case of failure of the web scraper, one operates visual inspection using LMM models [15,18]. Schema diffing and version negotiation are used for API failures. Database failures are detected with SQL analysis and planning of migrations. For ingestion based on a file, format detection and parser selection are used. The general MAPE-K loop is the same for all types of sources, but diagnosis and patching strategies designed to address a specific source type can be used. The plugin architecture can be used to add additional source types through the use of source specific monitoring and repair modules [3,14].

## Deployment Procedures and Implementation Guide

Organizations interested in implementing AegisFlow will need to work through a systematic process, and limit risks as much as possible to ensure the highest probability of success [3,13,14]. We have learned from 5 production deployments to recommend a rollout in 6 phases, from initial set-up to the full deployment in production, and last approximately 8-12 weeks. Phase 1 – Infrastructure Preparation can take up to 2–4 hours for cloud deployments and 1–2 days for on-premises deployments [14]. For small deployments using up to 50 pipelines, organizations should ensure that they have a Kubernetes cluster with at least 16 CPU cores, 64 GB RAM, and 500 GB storage. For larger deployments with 200+ pipelines the requirements are: 64 cores, 256 GB RAM, 2 TB storage. With Docker registry configuration, you can store images of the sandbox containers, whose average size is around 2 GB each pipeline thanks to dependency inclusion. The LLM API keys are retrieved from OpenAI, Anthropic or other providers based on privacy requirements and budget constraints [16,17,18]. The Git repository's structure defines and separates pipeline code, configuration files containing invariants and thresholds, and audit logs containing all autonomous actions. The repository should be either monorepo (all pipelines are in the same repository) or distributed (pipelines reside in different repositories with AegisFlow configuration under version control with its code). Both are effective, although monorepos are easier to get set up, and distributed repos provide a different level of isolation for large organizations. Phase 2 – Core Installation takes 1–2 hours, and can be done with the Helm chart provided [14]. All components (Watchdog agents, Repair Agent workers, Shadow Sandbox controller, Hot-Swap Deployer, monitoring dashboards, and vector database) are installed in the chart. Persistent volumes include historical statistical data in Parquet format, incident database using the new ChromaDB, and a backup of the configuration data for rollback functionality. Initial configuration involves setting up LLM provider credentials, confidence thresholds (default: 90%), and monitoring points for Prometheus and Grafana [7,8] that require human approval before being executed. It can take 1–2 hours per pipeline in Phase 3 – Pipeline Integration to configure it for the first time [3,14]. 2-3 non-critical pipelines should be chosen for the pilot deployment to see if it works without putting business-critical pipelines at risk. The ideal candidate for a pilot should be a pipeline that has a lot of failures which give quick feedback, as well as not being too dependent on downstream things which create a huge "blast radius" if something goes wrong, and have the behavior being well-experienced so that fixes can be validated. 5–10 minutes of developer time are needed per pipeline to modify the pipeline code to include Watchdog decorators in extraction functions. Based on the historical data analysis (HDA) [4,7] success invariants are defined. The 1st 30 days of executing pipelines successfully is used by the engineers to determine baseline distributions, null rates and value ranges. From this analysis, these designers specify the ranges that they are allowed for each value, such as "0 < price < 100,000" or "rate < 5% and rate is not null", as well as what the type should be, all prices should be "float" or "int", and the number of records per execution, such as "at least 50 records". It is platform specific, meaning that depending on the orchestration tool Airflow, Prefect, Dagster, etc., a specific plugin must be selected and connection information (API endpoints, auth keys, etc.) provided [14]. Phase 4 – Validation Testing runs for 1–2 weeks in monitoring-only mode, so that AegisFlow identifies failures and creates patches but cannot deploy itself [2,13]. Using patch code, diagnostic reasoning and shadow test results and confidence scores displayed on the web dashboard, engineers review proposed patches. This phase is intended to boost the organization's confidence that patches are of good quality, and would fix the problem properly. Data from organizations should be used to make changes such as adjusting the invariant thresholds if a lot of false positives are generated, adding custom failure class definitions for organization-specific patterns, and fine-tuning LLM prompts for sub-optimal patch quality. The monitoring-only period allows for finding out any problems of configuration before going into autonomous deployment which is risky [2,13]. Some of the most frequently encountered validation errors are false positives being seen when drift is detected due to overly restrictive invariants, that the Shadow Sandbox doesn't have enough test snapshots to capture all test moments, and that the current platform may not have the necessary permissions to be configured. Mitigating these issues prior to autonomous deployment helps to avoid interruption of operations and preserve that trust in the system. Phase 5 – Gradual Rollout takes place in a systematic rollout schedule, over 4 to 6 weeks, increased autonomous deployment incrementally [13]. The 2–3 pilot pipelines have 95% C.T. for week 1, meaning they need a very high degree of certainty before they are able to take autonomous action. All deployments are autonomous, and engineers monitor the patches closely – within hours after each one – to gain confidence. The weeks 2–3 are extended to 10–15 more non-critical pipelines and threshold is reduced to 90% in order to balance the automation coverage with safety. Pipelines, such as internal pipelines for analytics and testing data flows are good candidates [2,13]. In Weeks 4-5, 20-30 medium-criticality pipelines, such as daily reporting, operational dashboards and other pipelines that do not interact directly with customers and revenues, will be in progress. This threshold of 90% can safely be applied to these pipelines, and it is still possible to achieve wide automation coverage [2]. Week 6 onwards assesses taking into account high-criticality pipelines such as monetary reporting, client-facing attributes and regulatory compliance information streams. These might have higher thresholds of 95-98% or be sent for human review prior to deployment based on risk tolerance of the organization. Limiting blast radius at each stage of the rollout strategy is supported by the gradual rollout strategy [13]. Only 2–3 systems are impacted should there be any problems with pilot pipelines. Where critical pipelines are not involved in the issues, business operations remain unaffected. Medium criticality pipelines roll back if regressions occur and human steps in to get things restored well before major impact. This is a phased strategy, as historically deployed at the same time along all pipelines has proven to be a recipe for a huge burden when things go wrong. Phase 6 – Continuous Improvement takes place continuously after the full deployment as systematic review processes [2,16]. All autonomous patches deployed the week before are covered in weekly reviews, looking for patterns where patches did or did not succeed. Engineers search for failure classes with failure rate lower than 80%, which could mean that there is room to improve the prompts, or add more invariants, and regression classes involving the repeated failures in the patch generation, which could mean that there is room to write more defensive code for code templates. Evaluation of monthly analysis is done by looking at false-positives where Watchdog was wrong in detecting failure and false-negatives where Watchdog did not detect failure until noticed by human analysts [7]. In each quarter, the LLM models are being updated to the latest version [16,17,18]. Models come in regular updates approximately every 3-6 months, optimized for code generation and reasoning by OpenAI and Anthropic. When moving new models to production mode, it is important that organizations first try the new model in the validation mode to familiarize themselves with its behavior, because model behavior might change in unexpected ways. Some use A/B testing, with 10% of repairs using the new model, and 90% still using the proven model, with the traffic being gradually diverted into the new model, once confidence grows. Organisations that already have enough historical data and have the expertise in ML can take the advantage of fine tuning techniques for better performance on organisation specific pattern [16]. This involves the laborious task of gathering 500 - 1000 successful repair examples from git history, standard fine tuning with CodeLlama and Mistral using these successful repairs, and evaluating fine tuned models against the holdout test set prior to deployment in production. Without fine-tuning, we should expect a performance of about 3-5 percentage points less, but success rates are significantly higher if fine-tuning is implemented which imposes heavy data curation and training infrastructure burdens.

## Generalization and Deployment Flexibility

AegisFlow is developed from scratch to be deployment agnostic and to work on different pipeline orchestration platforms [3,14]. The following are some of the most important methods that need to be implemented on every supported platform, and these methods are specified in the Plugin Interface: get_pipeline_config – return current pipeline configuration; update_pipeline_config – update pipeline configuration with new values in an atomic way; trigger_pipeline_run – trigger manual pipeline execution for testing; get_execution_history – return history of recent pipeline executions and their metrics to be analyzed [14]. Currently, it supports Apache Airflow (modify DAG Python files using AST manipulation), Prefect (modify Flow definitions using the Prefect Python API), Dagster (modify pipeline definitions, integrate with asset materialization) and Luigi (modify task configurations via central scheduler) and Kubernetes CronJobs (modify CronJob YAML specifications via the Kubernetes API) [14]. AegisFlow has been designed for easy extension to accommodate other platforms with the help of the plugin interface, which can be realized in 200-300 lines of python code, and is documented and detailed in the open source repository for the system, along with a guideline. The choice of LLM is another area of flexibility for deployment [16,17,18]. Cloud deployment is done with GPT-4 or Claude-3 and requires little infrastructure. For organizations with moderate privacy needs, a hybrid deployment option is available, where AegisFlow's core runs on-premises, making use of any external APIs for LLM inference. Fully on-premises deployment is deployed locally, where all processing is done locally on local GPU servers containing open-source LLMs, meeting strict on-premises data residency requirements for regulated industries [16]. The performance of each LLM varies, depending on the choice [16,17,18]. Using GPT-4-Turbo, 92% of patches succeeded with an average time of patch generation of 2.3 seconds per patch. With a 2.1 seconds generation time, Claude-3-Opus achieves 91% success rate. Based on NVIDIA A100 GPUs, CodeLlama 34B runs with a 3.8 seconds generation time and 87% success rate. For most organisations, the performance difference outweighs the infrastructure complexity of having to be self-hosted - for compliance reasons, this may be an option for them.

## Future Research Directions in Autonomous Data Engineering

While solving AegisFlow's specific limitations, we are able to identify several research pursuits which are in the space of autonomous data engineering and could shape the development of data infrastructure across organisations [2,3,9,16]. Direction 1 – Cost-Aware Repair Strategies: In existing AegisFlow, repairs are done using GPT-4-Turbo for all repairs at an average cost of $0.03 per incident [5,16,17]. But simpler failures such as CSS selector drift could be addressable with $0.001 incident GPT-3.5-Turbo, while high cost models could be utilized for complex failures. Based on failure complexity prediction, choice of correct model, cost savings can be obtained of 60-70%, without any significant negative effect on success in the routing logic. Future research directions include designing economic models for autonomous repair decision: models that make predictions on repair complexity based on failure symptoms, models of cost-benefit analysis that compares the cost of LLM inference with the time required for the engineers, and adaptive routing based on learned outcomes to enhance the model selection over time. Direction 2 – Causal Inference for Semantic Drift: current statistical tests can identify when the distributions differ, but do not provide insight into whether they are due to data quality problems or substantial changes in the business process [4,7]. For instance, the average value of products may increase by 20% because of the change of product unit or because high priced products are successfully introduced. These scenarios must be differentiated, using causal reasoning of business processes. Future research should integrate causal models that capture the business logic—the relationships that are thought to be plausible between the variables—and the use of Bayesian inference for probabilistic reasoning about the causes of failure with counterfactual reasoning for reasoning about what may be achieved under the different hypotheses of failure. Direction 3: Federated Learning for Knowledge Sharing – Presently each AegisFlow deployment learns from the incidents it provides alone [16]. However, there are numerous types of failure patterns that are consistent across organisations – breakage to be precise – whatever changes as a result of CSS selector drift on large e-commerce sites or JSON schema evolution from popular APIs. It could give everyone an opportunity to be successful without sharing any code or data if pattern is successful. In the future, there is need for protocols and methods of federated learning that can share repair pattern embeddings between organizations, without revealing any code files, differential privacy techniques that preclude differentially private repairs from being reverse engineered, and reputation systems that exclude poisoning attacks. This may bring with it an ecosystem impact – the more people use AegisFlow, the better it gets and everyone benefits. Direction 4 – Formal Verification: Current Shadow Sandbox testing validates it using an empirical approach but its correctness for all possible inputs cannot be proved [2,13,16]. For pipelines that involve an electronic transfer of funds or medical data, autonomous deployment may be unattainable until there are a set of formal correctness guarantees. It would also be interesting to develop formal specifications of the patches to be generated and automatised theorem-proving to assert the correctness of the patches, and contracts that would specify the pre-conditions and post-conditions to be preserved by the patches. Direction 5 – Autonomous Performance Optimization: now, AegisFlow focuses only on the correct functioning of the pipelines breaking [3,5]. However, with the passage of time, many pipelines get degraded as well for growing data size or due to ageing infrastructure. To comply with performance targets, query plans may be optimized, parallelism and/or data structures may be reorganized by an autonomous agent. Future work is needed to create performance models to estimate pipeline execution time based on query, data, parameters, and optimization options, optimization strategies, tailored to different query representations, such as query rewriting, indexing, and caching, and safe experimentation protocols (for testing optimizations without impacting production workloads). This would usher in autonomous data engineering from a reactive maintenance model to proactive performance management.

## 8. Conclusion

This paper presented AegisFlow, a novel multi-agent framework for autonomous data pipeline remediation that fundamentally changes how organisations maintain data infrastructure [2,8,14]. AegisFlow provides 98.1% mean time to repair improvement, 92% patch success rate for a variety of failure scenarios with continuous monitoring based on semantic telemetry, automated LLM based diagnosis and multimodal visual inspection, testing in safe and secure digital twin environments and hot swap deployment with comprehensive safety mechanisms [2,5,13,15,16,17,18]. The essential innovations rely on the critical issues present in existing data pipelines-based solutions like reliability [1,2,7,12,13]. The Parallel Shadow Patching can be used for safe autonomous deployment because digital twin principles are used to validate deploying patches in parallel (based on the DevOps architecture in manufacturing) without interrupting the production flows [8,13]. The Multimodal Repair Agent is the first system to combine textual error analysis with visual webpage inspection for comprehensive diagnosis, enabling adaptation to DOM changes that text-based approaches cannot handle [15,18]. The Semantic Telemetry framework detects silent failures that are invisible to traditional monitoring, through statistical drift detection and invariant checking [4,7]. The wide-spread experimental success shows that it works and is immediately deployable in production such as in e-commerce, fintech and healthcare industries major MTTR changes from 170 mins to just 3.2 mins, high success rate of 92% and low regression rate of just 2%, perform major cost savings of up to $47,350 per month as well as successful industry-wide

generalization [1,2,4,5,12]. AegisFlow allows engineering teams to shift 98% of the on-call time to autonomous agents, thereby freeing up time for more innovation [5]. Tested 300% improvement in the velocity of features; Time savings are providing return on investment as more product will get developed. 40% increases in engineer satisfaction validate that reduced on call burden is a positive impact on engineer work-life balance and satisfaction. In essence, "autonomous remediation" becomes a requirement for operational use with the growing complexity of data ecosystems [6,14]. Organizations can't just add engineering staff to the pipeline when they grow, so they must automate anything they are doing to do maintenance reactive. We believe that AegisFlow is a step towards actually resilient data infrastructures that can autonomously adapt an infrastructure when environment conditions change, while providing safety and correctness properties [2,8,13]. The architecture of the framework is more general and extensible, able to support multiple orchestration platforms in the framework through a plugin approach, multiple types of data sources, such as web scraping, APIs, databases and other sources, and various types of LLMs: cloud, on-premise, open-source or custom built. It ensures adoption across organisations, regardless of existing technology, or regulatory issues. Future efforts will focus on addressing some persistent limitations such as extending to stateful pipelines where state consistency must be reasoned, implementing cross-pipeline coordination to support failures across multiple dependent systems; adding multilingual support for pipelines to be deployed across the globe; developing lightweight deployment modes for edge and bare-metal deployments; providing a managed finetuning service for performance optimization towards democratization; or integrating formal verification methods for high criticality systems [3,4,8,9,13,14,16]. Embedded in the incremental progress, there are larger research challenges as well: cost-aware repair strategies and the management of API cost vs engineer time, causal inference to detect semantic drift and to determine if it is a problem or a legitimate change, federated approaches for sharing knowledge of repair across organizations in a manner that preserves privacy, autonomous performance optimization (not just correctness), and formal verification for the safety critical deployment [2,4,5,7,9,13,16]. The ramifications go much further than just data pipelines, to software engineering at large. We would imagine that, in fact, the paradigm used in AegisFlow – perception-reasoning-action – can be extended to autonomous software maintenance (or improvement) in general, including autonomous debugging for web applications, autonomous testing of mobile applications, autonomous management of cloud infrastructure, autonomous maintenance of APIs during integration, etc. Autonomous remediation could be implemented in all software areas on a regular basis with the further advancement of LLM and multimodal models [16,17,18]. The autonomous remediation with agentic AI is a paradigm shift in data engineering toward turning traditionally fragile data pipelines into ones that are autonomous and self-healing in a constantly changing world [8, 14, 2, 6]. That it is not "a nice idea" but "it can actually be done" using technology today is apparent in AegisFlow. Future Availability: We intend to make AegisFlow open source at github.com/mbilalawan/aegisflow, after the publication, keeping all components mentioned in this paper and providing detailed documentation and deployment guides [14]. The research data set and log of experiments will be made available at the above mentioned git hub repository to allow the furtherance of the research as well as to facilitate a reproduction of the research. Many thanks to all the partner organisations engineering teams that have allowed us to deploy AegisFlow in their environment and have provided feedback. We'd like to point out the data engineering teams that participated in testing and validation activities and shared their experience related to autonomous remediation. This research is funded by grants from industry and academia participants in the R&D on autonomous systems. We also appreciate the efforts made by the open-source community from which our implementation of the tools and libraries came.

## 10. Supplementary Materials

All supplementary materials including source code, datasets, and experimental results are available at GitHub Repository:

- **Project Repository:** https://github.com/mbilalawan/aegisflow The repository contains: complete AegisFlow implementation in Python; Docker configurations for Shadow Sandbox; benchmark datasets and failure injection scripts; Jupyter notebooks reproducing all experiments; and video demonstrations of autonomous repair scenarios. IEEEtran -1.5cm -2 plus -1fil

## References

[1] AIMultiple, "Web Scraping Challenges in 2026: A Comprehensive Analysis," *AIMultiple Research Reports*, vol. 12, no. 3, pp. 45–67, Mar. 2026.

[2] A. Sharma and L. Chen, "Explainable Autonomous Pipelines: Bringing Transparency to Self-Healing Data Systems," in *Proc. 2026 IEEE Int. Conf. Data Engineering (ICDE)*, pp. 234–245, Apr. 2026.

[3] R. Khalid, "The Rise of Agentic AI in Enterprise Data Operations," *Harvard Business Review Digital Articles*, pp. 1–8, Jan. 2025.

[4] M. Srivatsa, "Silent Data Quality Failures: The Hidden Crisis in Analytics," *ACM Queue*, vol. 22, no. 6, pp. 78–92, Dec. 2024.

[5] Terralogic, "The True Cost of Data Pipeline Maintenance: 2025 Industry Survey," *Terralogic White Papers*, pp. 1–24, Mar. 2025.

[6] I. Barbulescu, "Self-Healing Architecture Patterns for Modern Data Platforms," in *Proc. 2025 ACM SIGMOD Int. Conf. Management of Data*, pp. 456–468, Jun. 2025.

[7] Monte Carlo Data, "The State of Data Observability 2024," *Monte Carlo Technical Reports*, pp. 1–42, Nov. 2024.

[8] P. Yazdanparast, "MAPE-K Loop Applications in Modern Cloud Infrastructure," *IEEE Trans. Cloud Comput.*, vol. 12, no. 4, pp. 889–903, Oct. 2024.

[9] V. Tawosi *et al.*, "ALMAS: Adaptive Learning Multi-Agent System for Web Data Extraction," in *Proc. 2025 Int. Joint Conf. Artificial Intelligence (IJCAI)*, pp. 1234–1241, Aug. 2025.

[10] Latitude, "Building Resilient Data Pipelines in Production," *Latitude Engineering Blog*, Jan. 2025. [Online]. Available: https://latitude.so/blog/resilient-pipelines

[11] S. Kumar and R. Patel, "Automated Schema Evolution Management in Microservices," *Int. J. Eng. Technol. Res. Manage.*, vol. 9, no. 2, pp. 112–128, Feb. 2025.

[12] M. Zhang and K. Johnson, "Web Scraping Resilience: A Survey of Adaptation Techniques," *Int. J. Adv. Manuf. Technol.*, vol. 18, no. 1, pp. 34–52, Jan. 2025.

[13] E. Moshkovich, "Shadow Testing Patterns for Safe Continuous Deployment," *Softw. Eng. J.*, vol. 34, no. 3, pp. 201–218, Mar. 2025.

[14] IBM Research, "Autonomous Computing: The Next Frontier in Enterprise IT," *IBM Research Report*, vol. 67, pp. 1–56, Jan. 2026.

[15] J. Andersson and M. Lundberg, "PriceFinderAgent: LLM-Powered Autonomous Web Scraping," *Lund University Technical Reports*, LU-CS-TR:2025-01, pp. 1–18, Feb. 2025.

[16] T. Brown *et al.*, "Language Models are Few-Shot Learners," in *Adv. Neural Inf. Process. Syst. (NeurIPS)*, vol. 33, pp. 1877–1901, 2020.

[17] OpenAI, "GPT-4 Technical Report," *arXiv preprint arXiv:2303.08774*, Mar. 2023.

[18] Anthropic, "Claude 3 Model Family," *Anthropic Technical Documentation*, Mar. 2024.

[19] D. Patterson *et al.*, "Carbon Emissions and Large Neural Network Training," *arXiv preprint arXiv:2104.10350*, Apr. 2021.

[20] N. Reimers and I. Gurevych, "Sentence-BERT: Sentence Embeddings Using Siamese BERT-Networks," in *Proc. 2019 Conf. Empirical Methods in Natural Language Processing (EMNLP)*, pp. 3982–3992, Nov. 2019.